\documentclass[10pt,twocolumn,letterpaper]{article}

\usepackage[pagenumbers]{cvpr} 

\usepackage[dvipsnames]{xcolor}

\definecolor{cvprblue}{rgb}{0.21,0.49,0.74}
\usepackage[pagebackref,breaklinks,colorlinks,citecolor=cvprblue]{hyperref}

\def\paperID{5098} 
\def\confName{CVPR}
\def\confYear{2024}

\title{Do You Guys Want to Dance: Zero-Shot Compositional Human Dance Generation with Multiple Persons}
\author{Zhe Xu, Kun Wei, Xu Yang, Cheng Deng\\
School of Electronic Engineering, Xidian University, Xi’an 710071, China\\
{zhexu@stu.xidian.edu.cn, \{chdeng.xd, xuyang.xd, weikunsk\}@gmail.com}}

\let\oldtwocolumn\twocolumn
\renewcommand\twocolumn[1][]{%
    \oldtwocolumn[{#1}{
    \begin{center}
           \includegraphics[width=1\textwidth]{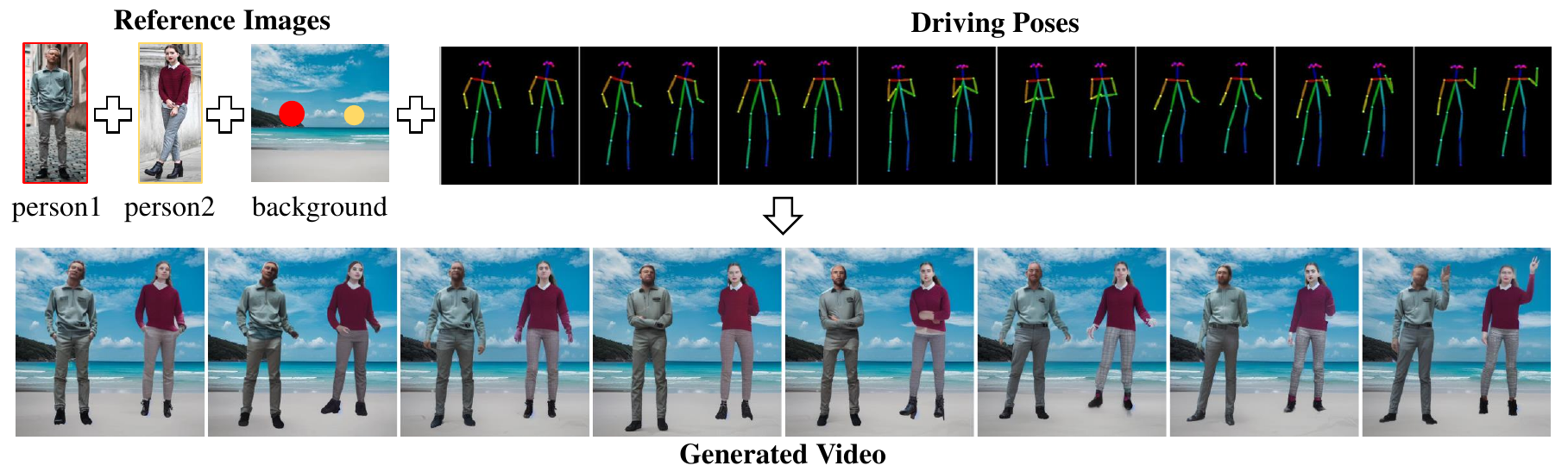}
           \captionof{figure}{(\emph{top}) Given reference images of multiple persons, background, and a sequence of driving poses, (\emph{bottom}) our method can synthesize a realistic video that contains all the persons in the background precisely following the poses without any video data utilized.}
           \label{task}
    \end{center}
    }]
}

\begin{document}
\maketitle

\begin{abstract}
Human dance generation (HDG) aims to synthesize realistic videos from images and sequences of driving poses. Despite great success, existing methods are limited to generating videos of a single person with specific backgrounds, while the generalizability for real-world scenarios with multiple persons and complex backgrounds remains unclear. To systematically measure the generalizability of HDG models, we introduce a new task, dataset, and evaluation protocol of compositional human dance generation (cHDG). Evaluating the state-of-the-art methods on cHDG, we empirically find that they fail to generalize to real-world scenarios. To tackle the issue, we propose a novel zero-shot framework, dubbed MultiDance-Zero, that can synthesize videos consistent with arbitrary multiple persons and background while precisely following the driving poses. Specifically, in contrast to straightforward DDIM or null-text inversion, we first present a pose-aware inversion method to obtain the noisy latent code and initialization text embeddings, which can accurately reconstruct the composed reference image. Since directly generating videos from them will lead to severe appearance inconsistency, we propose a compositional augmentation strategy to generate augmented images and utilize them to optimize a set of generalizable text embeddings. In addition, consistency-guided sampling is elaborated to encourage the background and keypoints of the estimated clean image at each reverse step to be close to those of the reference image, further improving the temporal consistency of generated videos. Extensive qualitative and quantitative results demonstrate the effectiveness and superiority of our approach.
\end{abstract}

\section{Introduction}
\indent\hspace{1em} Can you imagine you dancing on the street in front of your house? What about you and your friends dancing together on the beach you have wanted to go to for a long time? Behind these questions is a prominent computer vision task--human dance generation (HDG) \cite{wang2018video,wang2019few,siarohin2019first,chan2019everybody,karras2023dreampose,wang2023disco}, which aims to synthesize realistic videos from reference images and driving poses. Due to the promising applications in various fields such as entertainment and education, the past few years have witnessed an increasing interest and rapid improvement in HDG research.

The key challenge of the HDG task lies in simultaneously retaining the appearances of foreground and background consistent with the reference images while precisely following the driving poses. Existing methods tackle the challenge in a supervised manner, \emph{i.e.}, collecting numbers of video data and utilizing them as ground truth to train their models. Despite great success, several tricky issues remain: (i) they are limited to generating videos of a single person with specific backgrounds (\emph{e.g.}, fashion video with empty background \cite{karras2023dreampose} or TikTok short video \cite{wang2023disco}), the generalizability for real-world scenarios with multiple persons and complex backgrounds is not specifically tested. (ii) the number of required videos for training is large (\emph{e.g.}, about 500 different poses \cite{karras2023dreampose,wang2023disco} per person), which is time-consuming and expensive to obtain. 

\begin{figure}[]
\centering
\includegraphics[width=0.92\columnwidth]{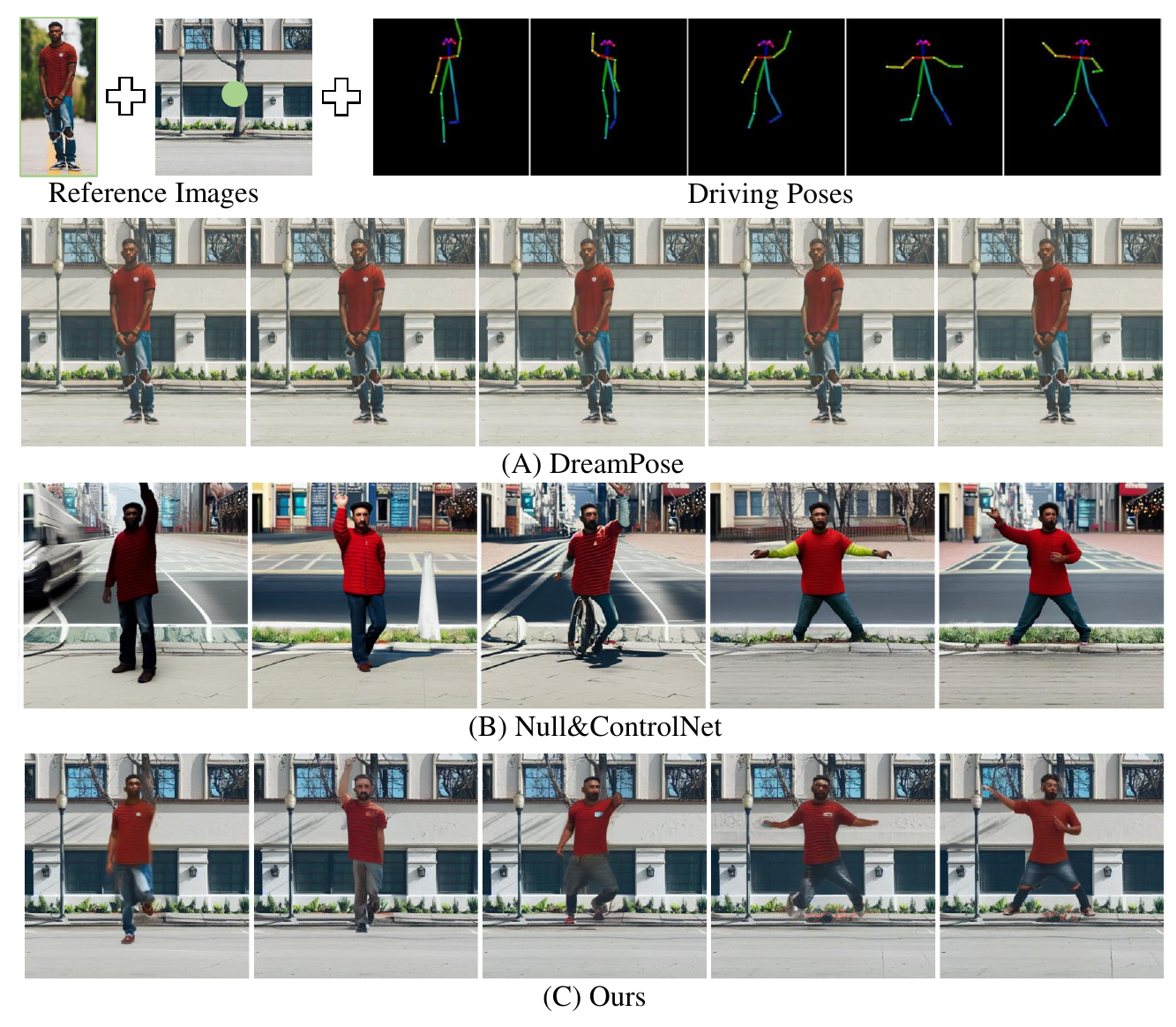} 
\caption{(\emph{top}) Given reference images and driving poses, (A) DreamPose \cite{karras2023dreampose}, a state-of-the-art HDG model, fails to generalize to real-world scenarios with complex background and diverse poses. (B) Directly combining null-text inversion \cite{mokady2023null} and ControlNet \cite{zhang2023adding} leads to severe temporal inconsistency. (C) Our approach can synthesize a video that simultaneously retains the appearance consistent with the reference images and precisely follows the driving poses.}
\label{motivation}
\end{figure}

To systematically measure the generalizability in real-world scenarios of existing models, we introduce a new task, compositional human dance generation (cHDG). Given reference images of multiple persons, background, and driving poses, cHDG aims to synthesize a video that simultaneously contains all the persons in the background precisely following the poses (Fig.\ref{task}). We also construct a new dataset that contains various persons, backgrounds, and pose sequences. Moreover, we propose a new evaluation protocol that measures temporal consistency and pose accuracy of the generated results. Evaluating the state-of-the-art methods on cHDG, we empirically find that they fail to generalize to the real-world scenarios (\emph{e.g.}, the second row of Fig.\ref{motivation}).  

For another thing, diffusion-based models (DMs) \cite{ho2020denoising,song2020denoising,nichol2021improved,croitoru2023diffusion,nichol2021glide,rombach2022high,saharia2022photorealistic} have achieved remarkable success in text-to-image generation and several methods \cite{avrahami2022blended,mokady2023null,clark2023text} are proposed to leverage pretrained text-to-image DMs for zero-shot image editing. Typically, these methods first invert the original real image to obtain the noisy latent code then denoise to generate the edited image under the target text condition. Inspired by this, given the composed image containing reference persons and background, our cHDG task can be treated as a pose-conditioned image-to-video editing problem. A naive solution to cHDG is inversion and then denoising via a pretrained pose-conditioned ControlNet \cite{zhang2023adding}. However, directly combining null-text inversion \cite{mokady2023null}, a state-of-the-art inversion method, and ControlNet leads to severe temporal inconsistency (the third row of Fig.\ref{motivation}), which can be attributed to the strong generative priors of the pretrained models.

In this paper, we propose MultiDance-Zero, a novel zero-shot framework for the challenging cHDG problem. Concretely, to accurately reconstruct the composed reference image, we first present a pose-aware inversion method to obtain the noisy latent code and initialization text embeddings using pretrained DMs. Despite near-perfect reconstruction for the reference image, directly generating videos from them conditioned on the driving poses will still lead to severe appearance shift. To this end, we propose to optimize a set of generalizable text embeddings that can generalize well from the noisy latent code of the reference image to unseen poses. This is achieved by utilizing the augmented images via a compositional augmentation strategy. Finally, we elaborate consistency-guided sampling to further improve the temporal consistency of generated videos. 

Our contributions can be summarized as three-fold:
\begin{itemize}
\item A new task, dataset, and evaluation protocol of cHDG, aiming to synthesize multi-person videos in real-world scenarios with complex backgrounds and diverse poses. 
\item A novel zero-shot approach consists of pose-aware inversion, generalizable text embeddings optimization, and consistency-guided sampling.
\item Extensive qualitative and quantitative results demonstrate the effectiveness and superiority of our approach.
\end{itemize}

\section{Related Work}

\subsection{Human Dance Generation}
\indent\hspace{1em} Given reference images and sequences of human body poses, human dance generation (HDG) aims to synthesize videos consistent with both reference images and driving motion. Early works \cite{siarohin2019first,siarohin2019animating,siarohin2021motion,wang2018video,wang2019few,zhao2022thin} require multiple separate networks for background prediction, motion representation, or occlusion map generation. Recently, DreamPose \cite{karras2023dreampose} proposes a diffusion-based method for generating fashion videos from still images, where a dual CLIP-VAE image encoder and an adapter module are employed to replace the text encoder in the original Stable Diffusion architecture. Meanwhile, a two-phase fine-tuning scheme is introduced to improve the image fidelity and temporal consistency. DisCo \cite{wang2023disco} presents a model architecture with disentangled foreground-background-pose control and human attribute pre-training to improve the fidelity of generation. Despite great success, they require a number of video data to train their models and fail to generalize to real-world scenarios with multiple persons and complex backgrounds. To tackle the challenges, we introduce a new cHDG task and propose a zero-shot approach.

\subsection{Diffusion-Based Editing}
\indent\hspace{1em} Diffusion models (DMs) \cite{ho2020denoising,song2020denoising} are a category of probabilistic generative models that learn to reverse a process that gradually degrades the training data structure and have achieved remarkable success in image and video generation \cite{croitoru2023diffusion,harvey2022flexible,rombach2022high,saharia2022photorealistic,esser2023structure,wu2023tune,ge2023preserve,yu2023video,luo2023videofusion,blattmann2023align,avrahami2023spatext}.
Due to theirs powerful generation ability, pretrained text-to-image DMs have been increasingly used for image and video editing \cite{avrahami2022blended,avrahami2023blended,kawar2023imagic,mokady2023null,cao2023masactrl,qi2023fatezero,tumanyan2023plug,balaji2022ediffi,yang2023paint,zhang2023sine}.
Blended Diffusion \cite{avrahami2022blended} performs region-based editing in natural images leveraging a pretrained CLIP model \cite{radford2021learning}.
Imagic \cite{kawar2023imagic} presents the first text-based semantic image editing technique that allows for complex non-rigid edits on a single real input image. Null-text inversion \cite{mokady2023null} proposes to optimize the null-text embeddings around a sequence of noisy pivot codes to invert real images into the latent space with impressive editing capabilities. MasaCtrl \cite{cao2023masactrl} develops a tuning-free method to achieve complex non-rigid image editing by a mutual self-attention mechanism.
FateZero \cite{qi2023fatezero} presents the first framework for zero-shot text-based video editing, where an Attention Blending Block is employed to fuse the attention maps in the inversion and generation process. 

Despite impressive progress, existing methods can only perform text-based editing that can be explicitly described by a target prompt. In our cases of cHDG, the reference image and target video frames share almost the same text prompt, \emph{e.g.}, ``a man and a woman standing on the beach". Applying the text-based editing methods to cHDG will lead to unsatisfying results especially for multiple persons, as shown later in Sec.\ref{exp:comparison}.

\subsection{Concept Customization}
\indent\hspace{1em} Another line of works related to our method is concept customization \cite{ruiz2023dreambooth,kumari2023multi,liu2023cones,gu2023mix,chefer2023attend,balaji2022ediffi,gal2022image}. DreamBooth \cite{ruiz2023dreambooth} first introduces the task of concept customization, which aims to generate a myriad of images of a subject in different contexts given some subject images. It fine-tunes a text-to-image DM with the input images paired with a text prompt containing a unique identifier.  Custom Diffusion \cite{kumari2023multi} proposes to fine-tune a small subset of model weights and can compose multiple subjects via closed-form constraints. Cones \cite{liu2023cones} locates the corresponding neurons that control the generation of the given subject by statistics of network gradients and uses those neurons to guide customized generation. Mix-of-Show \cite{gu2023mix} adopts an embedding-decomposed LoRA for single concept tuning and gradient fusion for center-node concept fusion. 

Our approach differs from the concept customization works in the following: (1) They are proposed for image generation with random poses. In contrast, we focus on human video generation that not only retains the appearance consistent with the reference images but also precisely follows the driving poses. (2) They require several different images for a single concept (\emph{e.g.} about 20 images per concept in Mix-of-Show \cite{gu2023mix}) to train their models, while only one image per concept is available in our case. (3) We do not tune the model weights, but they need to fine-tune the whole or part pretrained DMs, which is memory and computation inefficient. We include the comparisons in Sec.\ref{exp:comparison}.

\begin{figure*}[h]
\centering
\includegraphics[width=0.93\textwidth]{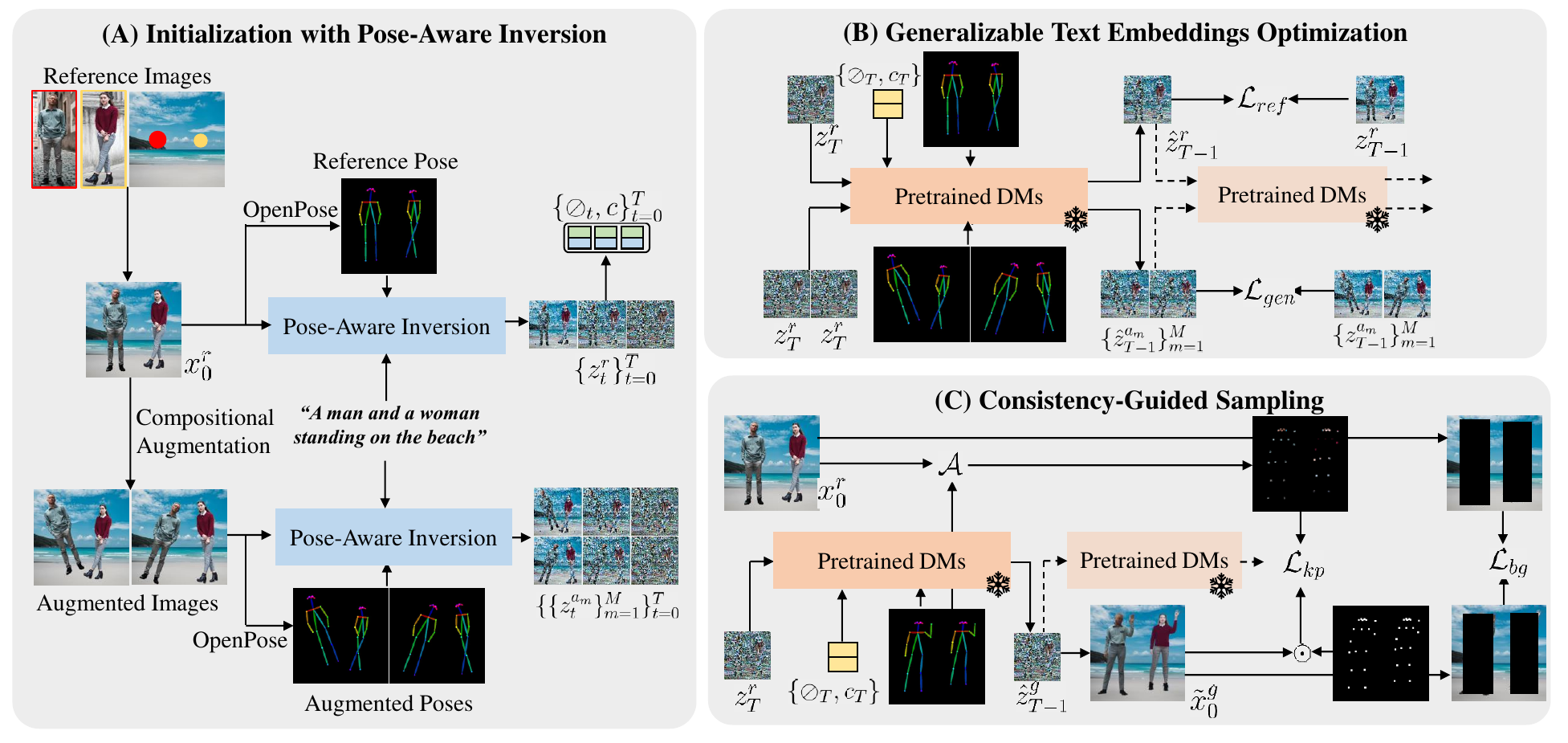} 
\caption{\textbf{Method overview.} Given reference images of multiple persons and background: (A) We propose pose-aware inversion to obtain the noisy latent code $z_T^r$ and initialization text embeddings from a composed reference image $x_0^r$ using pretrained DMs. Moreover, a compositional augmentation strategy is introduced to generate augmented images that share the same poses and appearances as $x_0^r$ but at different spatial locations. (B) We utilize the augmented images to optimize a set of generalizable text embeddings $\{ \oslash_t, c_t\}_{t=0}^T$, which is achieved by jointly minimizing a reference term $\mathcal{L}_{ref}$ and a generalization term $\mathcal{L}_{gen}$. (C) During inference, consistency-guided sampling is elaborated to encourage the background and keypoints of estimated clean image $\tilde{x}_0^g$ to be consistent with those of the reference image $x_0^r$, which can further improve the temporal consistency of generated videos. Red and yellow circles denote the user-provided location and scale of the corresponding persons.}
\label{framwork}
\end{figure*}


\section{Methodology}
\indent\hspace{1em} Given reference images of multiple persons, background, and a sequence of driving poses, our objective is to synthesize a realistic video that retains the appearance of foreground and background consistent with the reference images and precisely follows the driving poses. In Sec.\ref{subsec:preliminary}, we first provide some background on Latent Diffusion Model, ControlNet, DDIM inversion, and null-text inversion. Next, we present the details of our method in Sec.\ref{subsec:pa-inv}, Sec.\ref{subsec:text-opt}, and Sec.\ref{subsec:cg-sampling}. Fig.\ref{framwork} shows the overall framework of our method.

\subsection{Preliminaries}
\label{subsec:preliminary}

\textbf{Latent Diffusion Model} (LDM) \cite{rombach2022high} is proposed to perform diffusion and denoising in the latent space of a pretrained auto-encoder. Specifically, given an image $x_0$, an encoder $\mathcal{E}$ is first utilized to compress it to a low-resolution latent $z_0=\mathcal{E}(x_0)$, which can be reconstructed back to the image $\mathcal{D}(z_0) \approx x_0$ by decoder $\mathcal{D}$. Then, a UNet $\epsilon_{\theta}$ is trained to predict artificial noise following the objective:
\begin{equation}
    \min_{\theta} \mathbb{E}_{z_0, c, t, \epsilon \sim \mathcal{N}(0,I)} \Vert \epsilon-\epsilon_{\theta}(z_t,t,c) \Vert_2^2,
\end{equation}
where $c$ is the embedding of the text prompt and $z_t$ is a noisy sample obtained by adding noise to $z_0$ at timestamp $t$.

\noindent \textbf{ControlNet} \cite{zhang2023adding} is introduced to add spatial conditioning control, \emph{e.g.}, human pose in our case, to pretrained Latent Diffusion Model. Given spatial condition vector $p$ in the latent space, the training objective of ControlNet $\epsilon_{\phi}$ can be formulated as:
\begin{equation}
    \min_{\phi} \mathbb{E}_{z_0, c, p, t, \epsilon \sim \mathcal{N}(0,I)} \Vert \epsilon-\epsilon_{\theta}(z_t,t,c, \epsilon_{\phi}(z_t,t,c,p)) \Vert_2^2.
\end{equation}

\noindent \textbf{DDIM Inversion} is an inversion technique from the DDIM \cite{song2020denoising} sampling, based on the assumption that the ODE process can be reversed in the limit of small steps:
\begin{equation}
    z_{t+1} = \sqrt{\cfrac{\alpha_{t+1}}{\alpha_t}}z_t+(\sqrt{\cfrac{1}{\alpha_{t+1}}-1}-\sqrt{\cfrac{1}{\alpha_{t}}-1})\epsilon_{\theta}(z_t,t,c),
\end{equation}
where $\{\alpha_t\}_{t=0}^T$ are the schedule hyperparameters.

\noindent \textbf{Null-Text Inversion} \cite{mokady2023null} aims to invert real images with corresponding text prompts into the latent space of a text-guided diffusion model while maintaining its powerful editing capabilities.  It first performs DDIM inversion to compute a sequence of noisy codes, which roughly approximate the original image, then uses this sequence as a fixed pivot to optimize the input null-text embeddings.

\subsection{Pose-Aware Inversion}
\label{subsec:pa-inv}
\begin{figure}[]
\centering
\includegraphics[width=0.95\columnwidth]{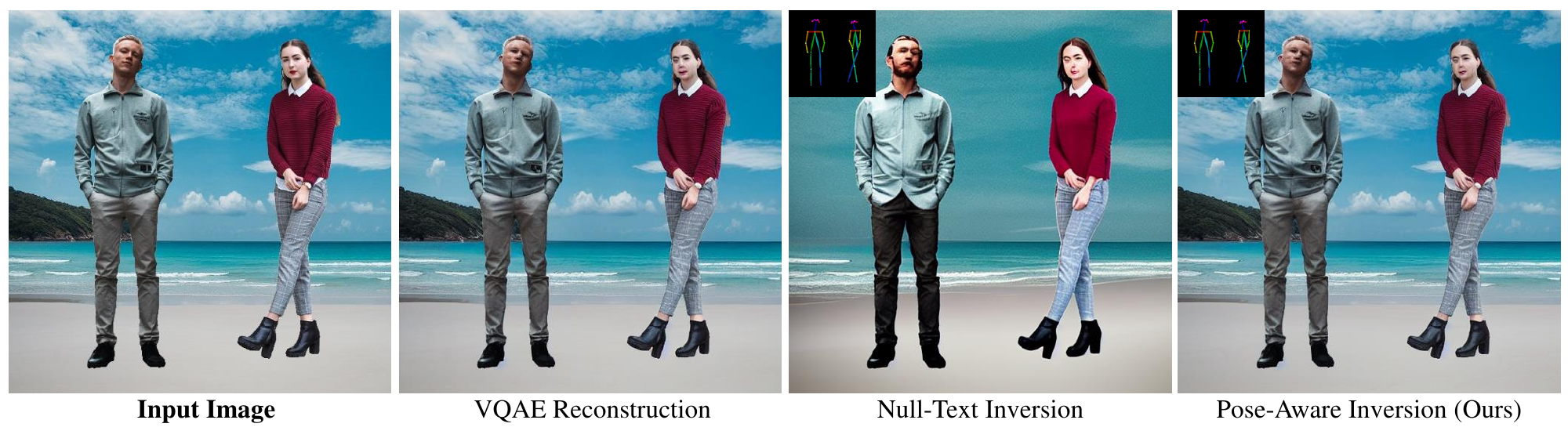} 
\caption{\textbf{Comparisons of reconstruction results}. We visually show the reconstruction results of the original null-text inversion and our pose-aware inversion. The pose inputs utilized in the reverse process are attached to the top left.}
\label{pa-inv}
\end{figure}

\indent\hspace{1em} Given a reference image $x_0^{r}$ composed of target persons and background, a naive solution to the cHDG problem is first performing null-text inversion to invert the image into the latent space and then denoising conditioned on the driving poses via a pretrained ControlNet. However, as shown in Fig.\ref{pa-inv}, original null-text inversion produces unsatisfying reconstruction when pose-conditioned ContolNet is applied. To this end, we introduce a pose-aware inversion method for accurate reconstruction with pose condition. Firstly, we generate a sequence of noisy codes $z_T^{r}, \cdots, z_0^{r}$ using a pose-aware variant of DDIM inversion:
\begin{equation}
\begin{aligned}
    z_{t+1}^r = &\sqrt{\cfrac{\alpha_{t+1}}{\alpha_t}}z_t^r +(\sqrt{\cfrac{1}{\alpha_{t+1}}-1}-\sqrt{\cfrac{1}{\alpha_{t}}-1}) \\ 
    &\epsilon_{\theta}(z_t^r,t,c,\epsilon_{\phi}(z_t^r,t,c,p^r)),
\end{aligned}
\end{equation}
where $\epsilon_{\theta}$ and $\epsilon_{\phi}$ represent pretrained LDM and ControlNet, respectively. $z_0^{r} = \mathcal{E}(x_0^r)$ is the latent encoding of reference image obtained by image encoder $\mathcal{E}$. $c$ and $p^r$ are the text embedding and reference pose vector, respectively.  

Then, we perform the following pose-aware optimization with default guidance scale $\omega =7.5$ for the timestamps $t=T, \cdots, 1$, each for $N$ iterations:
\begin{equation}
    \min_{\oslash_t} \Vert z_{t-1}^r - z_{t-1}(\hat{z}_t^r, \oslash_t, c, p^r) \Vert^2_2,
\end{equation}
where $\hat{z}_t$ represents the noisy latent at timestamp $t$ during optimization and $\hat{z}_T^r = z_T^r$. $z_{t-1}(\hat{z}_t, \oslash_t, c, p)$ denotes applying pose-aware DDIM sampling step using $\hat{z}_t$, the unconditional embedding $\oslash_t$, the text embedding $c$, and pose condition vector $p$, namely:
\begin{equation}
\label{eq: pa-ddim-sampling}
\begin{aligned}
    z_{t-1}&(\hat{z}_t, \oslash_t, c, p) = \sqrt{\cfrac{\alpha_{t-1}}{\alpha_t}} \hat{z}_t \\
    & +(\sqrt{\cfrac{1}{\alpha_{t-1}}-1}-\sqrt{\cfrac{1}{\alpha_{t}}-1}) \hat{\epsilon}_{\theta}(\hat{z}_t, \oslash_t, c, p),
\end{aligned}
\end{equation}
where
\begin{equation}
\begin{aligned}
    \hat{\epsilon}_{\theta}(\hat{z}_t, \oslash_t, c, p) &= \omega \cdot \epsilon_{\theta}(\hat{z}_t,t,c,\epsilon_{\phi}(\hat{z}_t,t,c,p))\\
    &+ (1-\omega) \cdot \epsilon_{\theta} (\hat{z}_t,t,\oslash_t,\epsilon_{\phi}(\hat{z}_t,t,\oslash_t,p)).
\end{aligned}
\end{equation}

At the end of each step, we update 
\begin{equation}
\label{eq: pa-update}
    \hat{z}_{t-1} =  z_{t-1}(\hat{z}_t, \oslash_t, c, p).
\end{equation}

After optimization, we obtain a noisy latent $z_T^r$ and a set of null-text embeddings $\{ \oslash_t \}_{t=1}^T$. As shown in Fig.\ref{pa-inv}, they can be utilized to accurately reconstruct the reference image via Eq.\ref{eq: pa-ddim-sampling} and Eq.\ref{eq: pa-update}.

\subsection{Generalizable Text Embeddings Optimization}
\label{subsec:text-opt}
\indent\hspace{1em} Although the noisy latent $z_T^r$ and optimized null-text embeddings $\{ \oslash_t \}_{t=1}^T$ can accurately reconstruct the reference image, severe appearance inconsistency is observed when the reference pose condition is replaced by a target one. We attribute this to the strong pose prior within pretrained ControlNet. It is a challenging problem especially in the zero-shot setting with no additional images available.
In this paper, we introduce a compositional augmentation strategy to generate several augmented images and utilize them to optimize a set of generalizable text embeddings. 

Specifically, given reference images of multiple persons and background, we randomly scale, rotate, and displace the persons and compose with the background to obtain the augmented images $\{ x_0^{a_m}\}_{m=1}^M$ and pose vectors $\{p^{a_m}\}_{m=1}^M$, where $M$ is the number of augmented images (see Fig.\ref{framwork}). The persons in the augmented images share the same poses and appearances as those in the reference image but at different spatial locations, which can be used for learning generalizable text embeddings.

Then, sequences of noisy codes $\{ z_T^{a_m}, \cdots, z_0^{a_m}\}_{m=1}^M$ are generated by the pose-aware DDIM inversion:
\begin{equation}
\begin{aligned}
    z_{t+1}^{a_m} = &\sqrt{\cfrac{\alpha_{t+1}}{\alpha_t}}z_t^{a_m} +(\sqrt{\cfrac{1}{\alpha_{t+1}}-1}-\sqrt{\cfrac{1}{\alpha_{t}}-1}) \\ 
    &\epsilon_{\theta}(z_t^{a_m},t,c,\epsilon_{\phi}(z_t^{a_m},t,c,p^{a_m})),
\end{aligned}
\end{equation}
where $z_0^{a_m} = \mathcal{E} (x_0^{a_m})$ is the encoding of the $x_0^{a_m}$.

Unlike text-based editing with different source and target text prompts, the reference image and target video frames share the same text prompt in our case. Therefore, we optimize both the unconditional and conditional text embeddings using the following objective:
\begin{equation}
\label{equ:text-opt}
\begin{aligned}
    &\min_{\oslash_t, c_t} \underbrace{\Vert z_{t-1}^r - z_{t-1}(\hat{z}_t^r, \oslash_t, c_t, p^r) \Vert^2_2}_{\operatorname{reference} \operatorname{term} \mathcal{L}_{ref}}\\ 
    & + \underbrace{ \cfrac{\lambda_1}{M} \sum \nolimits_{m=1}^M \Vert z_{t-1}^{a_m} - z_{t-1}(\hat{z}_t^{a_m}, \oslash_t, c_t, p^{a_m}) \Vert^2_2}_{\operatorname{generalization} \operatorname{term}\mathcal{L}_{gen}},
\end{aligned}
\end{equation}
where $\{ \oslash_t, c_t \}$ is the generalizable text embeddings at step $t$. The first and second terms are reference and generalization terms, respectively.  The reference term encourages the learned text embeddings to accurately reconstruct the reference image while the generalization term enforces them to generalize to the augmented poses. $\lambda_1$ is the weighting coefficient. Notably, the start points of optimization for both the reference and generalization terms are the noisy latent from the reference image, \emph{i.e.}, $\hat{z}_T^r = \hat{z}_T^{a_m} = z_T^r$.

\begin{figure*}[]
\centering
\includegraphics[width=0.83\textwidth]{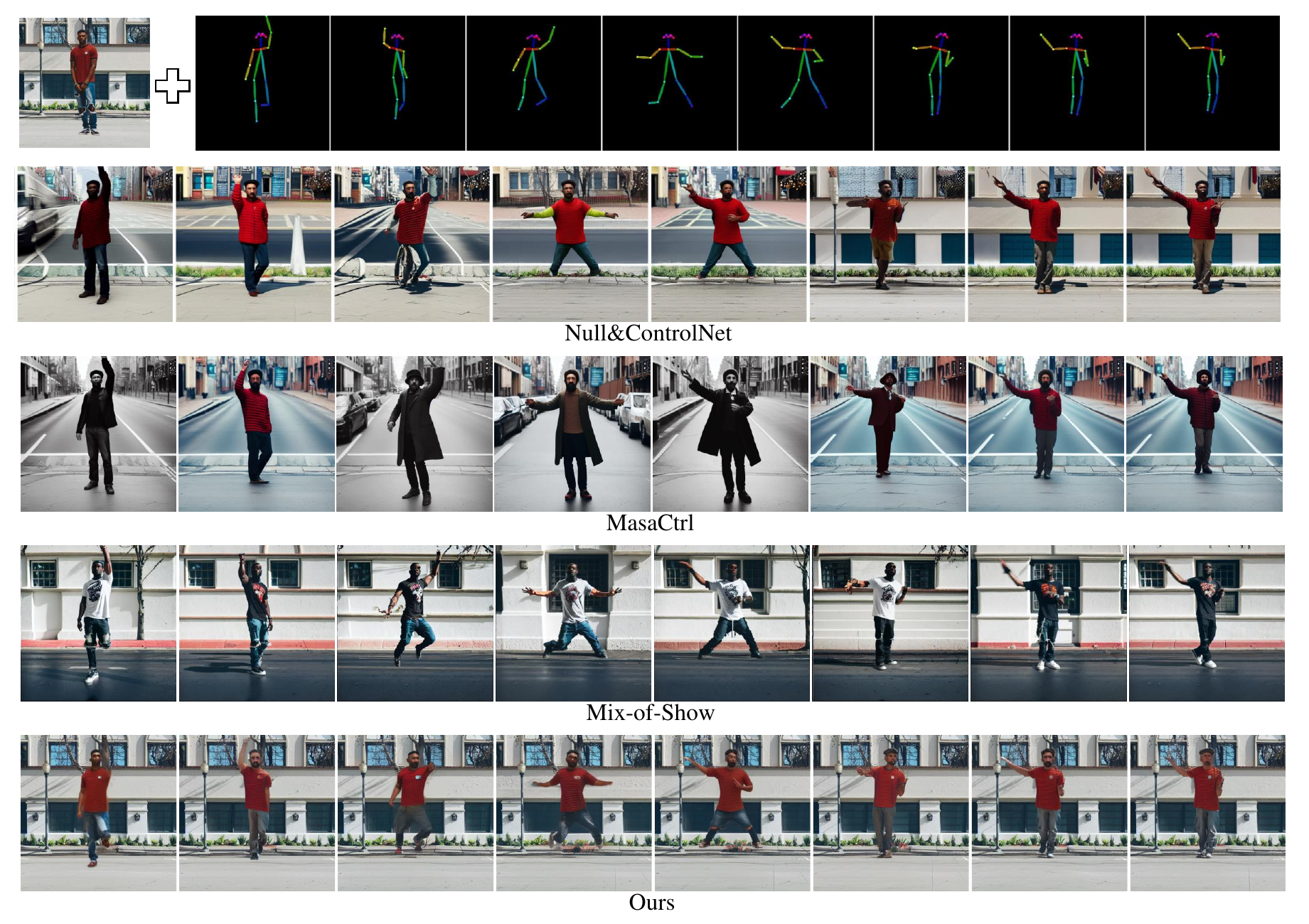} 
\caption{\textbf{Single-person results.} We compare our method with state-of-the-art baselines on cHDG with a single person. The top row shows the composed reference image and driving poses.}
\label{fig:comparison_1}
\end{figure*}

\begin{figure*}[]
\centering
\includegraphics[width=0.83\textwidth]{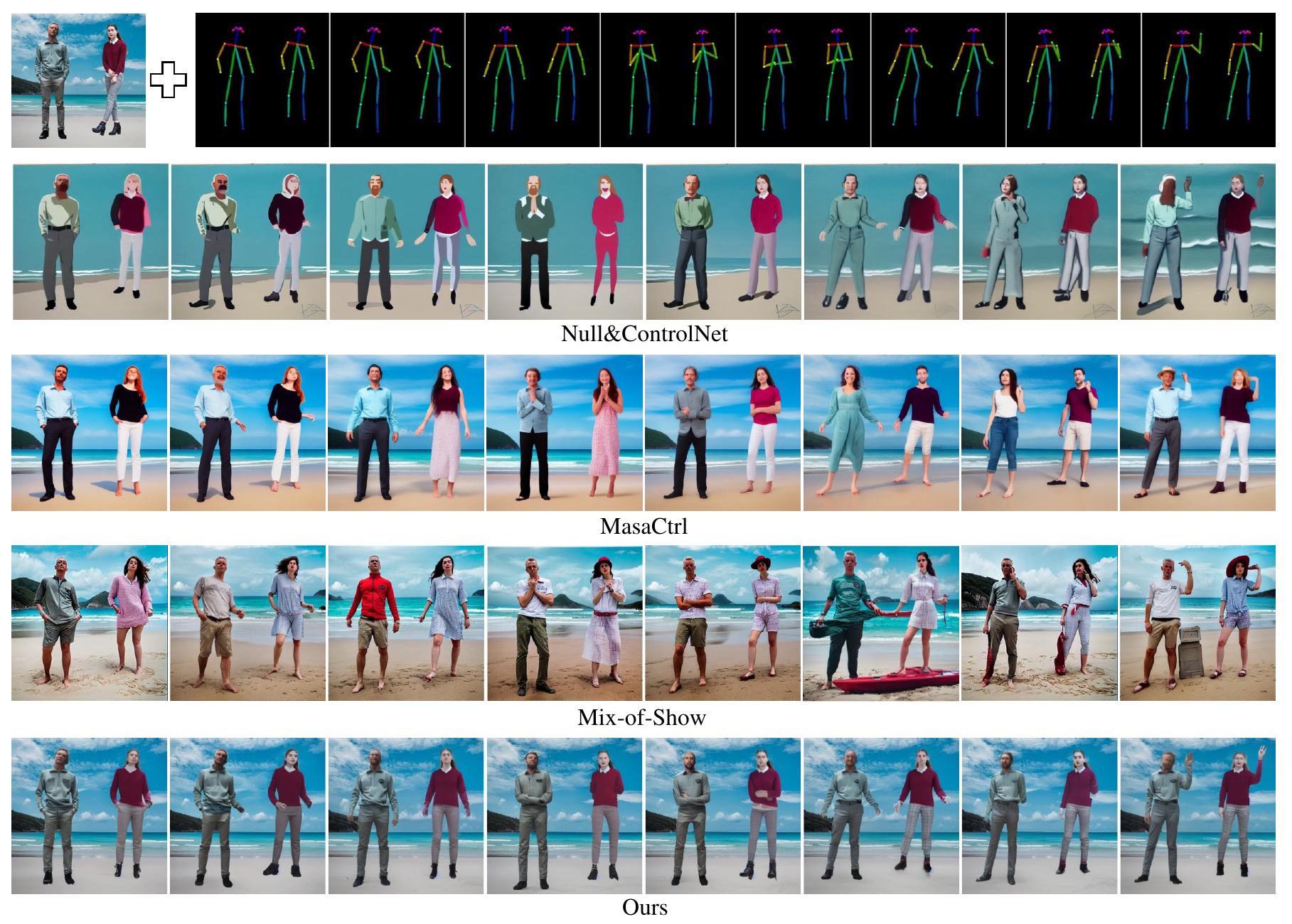} 
\caption{\textbf{Two-person results.} We compare our method with state-of-the-art baselines on cHDG with two persons. }
\label{fig:comparison_2}
\end{figure*}

\begin{figure*}[]
\centering
\includegraphics[width=0.83\textwidth]{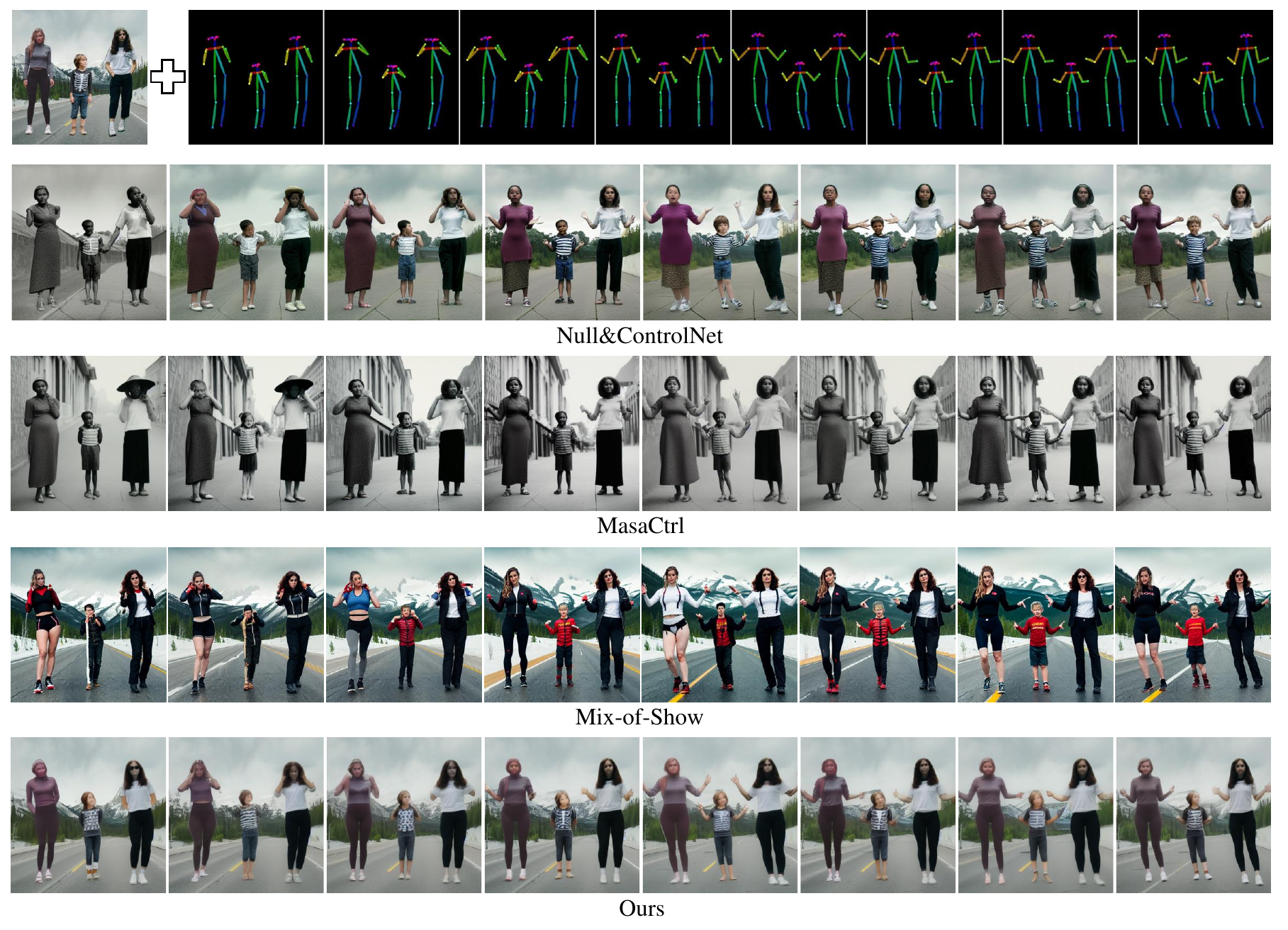} 
\caption{\textbf{Three-person results.} We compare our method with state-of-the-art baselines on cHDG with three persons.}
\label{fig:comparison_3}
\end{figure*}

\subsection{Consistency-Guided Sampling}
\label{subsec:cg-sampling}
\indent\hspace{1em} After optimization, the generalizable text-embeddings $\{ \oslash_t, c_t \}_{t=1}^T$, together with the noisy latent code $z_T^r$, can largely improve the consistency of generated videos with the reference images. However, since there is a gap between the augmented images and real images with diverse poses, some easily visible inconsistency still exists. To this end, we present a consistency-guided sampling method to further improve the temporal consistency of generated videos.

Inspired by recent guidance methods \cite{chung2022improving,chung2022diffusion} for conditional generation, we first estimate clean image $\tilde{x}_0^g$ from the sample $z_t^g$ using the Tweedie's formula \cite{efron2011tweedie}:
\begin{equation}
\begin{aligned}
    &\tilde{z}_0^g = \cfrac{z_t^g}{\sqrt{\bar{\alpha}_t}} - \cfrac{\sqrt{1-\bar{\alpha}_t}}{\sqrt{\bar{\alpha}_t}} \epsilon_{\theta}(z_t^g,t,c,\epsilon_{\phi}(z_t^g,t,c,p^g)),\\
    &\tilde{x}_0^g = \mathcal{D}(\tilde{z}_0^g),
\end{aligned}
\end{equation}
where $p^g$ and $\mathcal{D}$ are the target pose vector and the decoder of LDM, respectively. $\bar{\alpha}_t := \prod \nolimits_{s=0}^t \alpha_s$ and $z_T^g = z_T^r$.

Then, the sampling from the reverse diffusion can be formulated as:
\begin{equation}
\label{equ:sampling}
\begin{aligned}
    &z'_{t-1} = \cfrac{1}{\sqrt{\alpha_t}}(z_t^g-\cfrac{1-\alpha_t}{\sqrt{1-\bar{\alpha}_t}} \epsilon_{\theta}(z_t^g,t,c,\epsilon_{\phi}(z_t^g,t,c,p^g))),\\
    &z_{t-1}^g = z'_{t-1} - \delta_{t-1} \nabla_{z_t^g}\mathcal{L}(\tilde{x}_0^g),
\end{aligned}
\end{equation}
where $\delta_{t-1} = \delta / \mathcal{L}(\tilde{x}_0^g)$ is the dynamic step size inversely proportional to $\mathcal{L}(\tilde{x}_0^g)$. $\mathcal{L}(\tilde{x}_0^g) = \lambda_2 \mathcal{L}_{bg}(\tilde{x}_0^g) + \lambda_3 \mathcal{L}_{kp}(\tilde{x}_0^g)$ is the cost function with respect to estimated clean image, which consists of a background term and a keypoint term. Concretely, the background term measures the distance between the background of the estimated clean image and that of the reference image. And the keypoint term denotes the difference between the the keypoint (\emph{i.e.}, 18 keypoints of human sketch detected via OpenPose \cite{cao2017realtime}) values of the estimated clean image and that of the reference image:
\begin{equation}
\begin{aligned}
    &\mathcal{L}_{bg}(\tilde{x}_0^g) = \Vert x_0^r \odot m_{bg} - \bar{x}_0^g \odot m_{bg} \Vert_2^2,\\
    &\mathcal{L}_{kp}(\tilde{x}_0^g) = \Vert \mathcal{A}(x_0^r,p^g) - \bar{x}_0^g \odot m_{kp} \Vert_2^2,
\end{aligned}
\end{equation}
where $m_{bg}$ and $m_{kp}$ represent the background and keypoint masks, respectively. $\odot$ denotes element-wise multiplication. $\mathcal{A}$ is a function that assigns keypoints value for the target pose according to the reference image (see Fig.\ref{framwork}).

\section{Experiments}

\subsection{Dataset and Evaluation}
\textbf{Dataset.}
We collected a dataset consisting of 10 persons, 10 backgrounds, and 10 pose sequences, which can be composed to synthesize thousands of different videos. Images of the persons and backgrounds were sourced from Unsplash \cite{unsplash}, covering a wide range of genders, ages, clothes of persons and various scenes such as beach, snow, park, room, and mountain. Poses sequences were computed via OpenPose \cite{cao2017realtime} using videos from the public TikTok dataset \cite{jafarian2021learning}. 
For the evaluation, we randomly compose the persons, backgrounds, and pose sequences to synthesize 10 videos. Additional details of the dataset can be found in the \textbf{supplementary material}.

\noindent \textbf{Evaluation Metrics.}
Since there is no previous works on zero-shot cHDG, we propose to utilize the following metrics for evaluation borrowed from related tasks \cite{qi2023fatezero,ruiz2023dreambooth,cao2017realtime,xian2017zero,romera2015embarrassingly}: 
(1) `CLIP-I': the average cosine similarities between the composed reference image and each frame via pretrained a CLIP \cite{radford2021learning} model.
(2) `DINO':  the average pairwise cosine similarity between the DINO \cite{caron2021emerging} embeddings of generated frames and those of the composed reference image. 
(3) `mAP': we take the input pose sequences as ground truth and compute the mean average precision of generated videos to measure the pose accuracy.
(4) `H': harmonic mean value of DINO score and mAP to simultaneously measure the temporal consistency and pose accuracy, \emph{i.e.}, $\operatorname{H} = \cfrac{2\times \operatorname{DINO} \times \operatorname{mAP}}{\operatorname{DINO} + \operatorname{mAP}}$.

\subsection{Implementation Details}
\indent\hspace{1em} We use publicly available Stable Diffusion v1.5 \cite{rombach2022high} and pose-conditioned ControlNet \cite{zhang2023adding} as the base models. The inversion and sampling step $T$ is set to 50. For initialization with pose-aware inversion, we optimize the null-text embedding with a learning rate 0.01 and 50 iterations per step. For generalizable text embedding optimization, we utilize SAM \cite{kirillov2023segment} to separate the human foreground and set the number of augmented images, the learning rate, and the iterations per step to 16, 0.1, and 100, respectively. The weighting coefficients $\lambda_1$ in Eq.\ref{equ:text-opt}, $\delta$, $\lambda_2$ and $\lambda_3$ in Eq.\ref{equ:sampling} are empirically set to 1, 10, 100, and 2000, respectively.

\begin{table}[]
\centering
\begin{tabular}{p{2.3cm}p{1.2cm}p{1cm}p{1cm}p{0.8cm}}
\hline\noalign{\smallskip}
Method  &CLIP-I$\uparrow$ &DINO$\uparrow$  &mAP$\uparrow$ &H$\uparrow$\\
\hline\noalign{\smallskip}
Null\&ControlNet  &0.69 &0.61 &0.87 &0.72\\
MasaCtrl&0.65 &0.51 &0.93 &0.66\\
Mix-of-Show&0.77& 0.65 &\textbf{0.97} &0.78\\
\hline\noalign{\smallskip}
Ours, PAI &0.75&0.66&0.81 &0.73\\
Ours, GTEO &0.82&0.81&0.91 &0.85\\
Ours, Full &\textbf{0.83}&\textbf{0.83}&0.91 &\textbf{0.87}\\
\hline\noalign{\smallskip}
\end{tabular}
\caption{\textbf{Quantitative results.} We report four metrics for compared baselines and different variants of our approach.}
\label{tab:comparison}
\end{table}

\subsection{Comparisons}
\label{exp:comparison}
\indent\hspace{1em} Since there are no available zero-shot methods for cHDG, we build the following state-of-the-art baselines for comparison: (1) Null\&ControlNet first uses null-text inversion \cite{mokady2023null} to invert the reference image into the latent space and optimize a set of null-text embeddings, then generates videos via pretrained pose-conditioned ControlNet \cite{zhang2023adding}. (2) MasaCtrl \cite{cao2023masactrl} performs complex non-rigid frame-wise image editing via a mutual self-attention mechanism. (3) Mix-of-Show \cite{gu2023mix} adopts an embedding-decomposed LoRA for single concept tuning and gradient fusion for center-node concept fusion. We use the same text prompts and base models for fair comparison. Due to space limitations, we provide additional details and comparisons with other state-of-the-art methods in the \textbf{supplementary material}.

We present the quantitative results in Tab.\ref{tab:comparison}. As indicated by the harmonic mean value H, our method achieves the best temporal consistency and pose accuracy against baselines.
Fig.\ref{fig:comparison_1}, Fig.\ref{fig:comparison_2}, and Fig.\ref{fig:comparison_3}  show the qualitative results with different number of persons, which further verifies the superiority of our method.

\subsection{Ablation Studies}

\begin{figure}[]
\centering
\includegraphics[width=0.9\columnwidth]{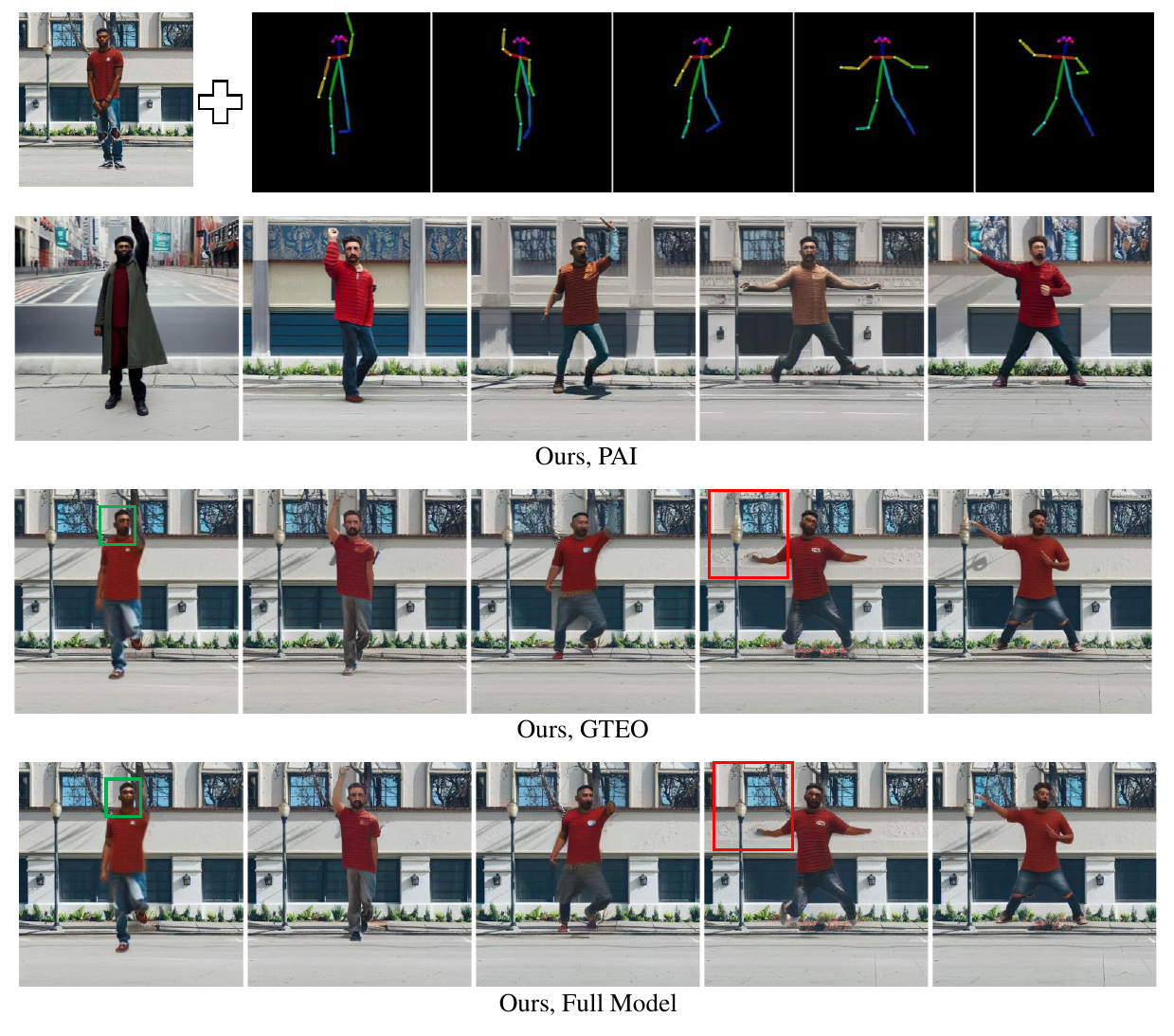} 
\caption{\textbf{Ablation study.} We present qualitative results of three variants of our method.}
\label{fig:ablation}
\end{figure}

\indent\hspace{1em} We perform ablation studies to verify the effectiveness of each component of our method. Specifically, three variants of our methods are compared: (1) Pose-Aware Inversion (PAI), (2) Generalizable Text Embedding Optimization (GTEO), and (3) Full Model. The quantitative and qualitative results are presented in Tab.\ref{tab:comparison} and Fig.\ref{fig:ablation}, respectively.
Although pose-aware inversion can accurately reconstruct the reference image, severe appearance inconsistency still exists when applied to the driving poses. 
Generalizable text embeddings optimization can generalize much better to unseen poses and our full model with consistency-guided sampling can further improve the temporal consistency.

\section{Conclusion and Discussion}
\indent\hspace{1em} In this paper, we introduce a new task, dataset, and evaluation protocol of cHDG, which aims to synthesize realistic videos consistent with reference images of multiple persons and complex backgrounds while precisely following the driving poses. We propose a novel zero-shot framework, MultiDance-Zero, for this challenging task. A pose-aware inversion method is first presented to obtain the initialization noisy latent code and text embeddings using pretrained DMs. We then propose to optimize a set of generalizable text embeddings using augmented images obtained by a compositional augmentation strategy. Moreover, consistency-guided sampling is elaborated to further improve the temporal consistency of generated videos. Extensive experiments verify the effectiveness and superiority over state-of-the-art baselines of our approach.

Our method inherits the limitations of pretrained DMs, \emph{e.g.},  the information
lost in the VQAE (see Fig.\ref{pa-inv}) affects the generation of high-quality facial and hand details. Using pixel-based DMs such as Imagen\cite{saharia2022photorealistic} or fine-tuning the VQAE may provide potential solutions and we leave this for future work.

{
    \small
    \bibliographystyle{ieeenat_fullname}
    \bibliography{main}

\begin{thebibliography}{53}
\providecommand{\natexlab}[1]{#1}
\providecommand{\url}[1]{\texttt{#1}}
\expandafter\ifx\csname urlstyle\endcsname\relax
  \providecommand{\doi}[1]{doi: #1}\else
  \providecommand{\doi}{doi: \begingroup \urlstyle{rm}\Url}\fi

\bibitem[Avrahami et~al.(2022)Avrahami, Lischinski, and Fried]{avrahami2022blended}
Omri Avrahami, Dani Lischinski, and Ohad Fried.
\newblock Blended diffusion for text-driven editing of natural images.
\newblock In \emph{CVPR}, pages 18208--18218, 2022.

\bibitem[Avrahami et~al.(2023{\natexlab{a}})Avrahami, Fried, and Lischinski]{avrahami2023blended}
Omri Avrahami, Ohad Fried, and Dani Lischinski.
\newblock Blended latent diffusion.
\newblock \emph{ACM Transactions on Graphics (TOG)}, 42\penalty0 (4):\penalty0 1--11, 2023{\natexlab{a}}.

\bibitem[Avrahami et~al.(2023{\natexlab{b}})Avrahami, Hayes, Gafni, Gupta, Taigman, Parikh, Lischinski, Fried, and Yin]{avrahami2023spatext}
Omri Avrahami, Thomas Hayes, Oran Gafni, Sonal Gupta, Yaniv Taigman, Devi Parikh, Dani Lischinski, Ohad Fried, and Xi Yin.
\newblock Spatext: Spatio-textual representation for controllable image generation.
\newblock In \emph{CVPR}, pages 18370--18380, 2023{\natexlab{b}}.

\bibitem[Balaji et~al.(2022)Balaji, Nah, Huang, Vahdat, Song, Kreis, Aittala, Aila, Laine, Catanzaro, et~al.]{balaji2022ediffi}
Yogesh Balaji, Seungjun Nah, Xun Huang, Arash Vahdat, Jiaming Song, Karsten Kreis, Miika Aittala, Timo Aila, Samuli Laine, Bryan Catanzaro, et~al.
\newblock ediffi: Text-to-image diffusion models with an ensemble of expert denoisers.
\newblock \emph{arXiv preprint arXiv:2211.01324}, 2022.

\bibitem[Blattmann et~al.(2023)Blattmann, Rombach, Ling, Dockhorn, Kim, Fidler, and Kreis]{blattmann2023align}
Andreas Blattmann, Robin Rombach, Huan Ling, Tim Dockhorn, Seung~Wook Kim, Sanja Fidler, and Karsten Kreis.
\newblock Align your latents: High-resolution video synthesis with latent diffusion models.
\newblock In \emph{CVPR}, pages 22563--22575, 2023.

\bibitem[Cao et~al.(2023)Cao, Wang, Qi, Shan, Qie, and Zheng]{cao2023masactrl}
Mingdeng Cao, Xintao Wang, Zhongang Qi, Ying Shan, Xiaohu Qie, and Yinqiang Zheng.
\newblock Masactrl: Tuning-free mutual self-attention control for consistent image synthesis and editing.
\newblock \emph{arXiv preprint arXiv:2304.08465}, 2023.

\bibitem[Cao et~al.(2017)Cao, Simon, Wei, and Sheikh]{cao2017realtime}
Zhe Cao, Tomas Simon, Shih-En Wei, and Yaser Sheikh.
\newblock Realtime multi-person 2d pose estimation using part affinity fields.
\newblock In \emph{CVPR}, pages 7291--7299, 2017.

\bibitem[Caron et~al.(2021)Caron, Touvron, Misra, J{\'e}gou, Mairal, Bojanowski, and Joulin]{caron2021emerging}
Mathilde Caron, Hugo Touvron, Ishan Misra, Herv{\'e} J{\'e}gou, Julien Mairal, Piotr Bojanowski, and Armand Joulin.
\newblock Emerging properties in self-supervised vision transformers.
\newblock In \emph{ICCV}, pages 9650--9660, 2021.

\bibitem[Chan et~al.(2019)Chan, Ginosar, Zhou, and Efros]{chan2019everybody}
Caroline Chan, Shiry Ginosar, Tinghui Zhou, and Alexei~A Efros.
\newblock Everybody dance now.
\newblock In \emph{ICCV}, pages 5933--5942, 2019.

\bibitem[Chefer et~al.(2023)Chefer, Alaluf, Vinker, Wolf, and Cohen-Or]{chefer2023attend}
Hila Chefer, Yuval Alaluf, Yael Vinker, Lior Wolf, and Daniel Cohen-Or.
\newblock Attend-and-excite: Attention-based semantic guidance for text-to-image diffusion models.
\newblock \emph{ACM Transactions on Graphics (TOG)}, 42\penalty0 (4):\penalty0 1--10, 2023.

\bibitem[Chung et~al.(2022{\natexlab{a}})Chung, Kim, Mccann, Klasky, and Ye]{chung2022diffusion}
Hyungjin Chung, Jeongsol Kim, Michael~T Mccann, Marc~L Klasky, and Jong~Chul Ye.
\newblock Diffusion posterior sampling for general noisy inverse problems.
\newblock \emph{arXiv preprint arXiv:2209.14687}, 2022{\natexlab{a}}.

\bibitem[Chung et~al.(2022{\natexlab{b}})Chung, Sim, Ryu, and Ye]{chung2022improving}
Hyungjin Chung, Byeongsu Sim, Dohoon Ryu, and Jong~Chul Ye.
\newblock Improving diffusion models for inverse problems using manifold constraints.
\newblock \emph{NeurIPS}, 35:\penalty0 25683--25696, 2022{\natexlab{b}}.

\bibitem[Clark and Jaini(2023)]{clark2023text}
Kevin Clark and Priyank Jaini.
\newblock Text-to-image diffusion models are zero-shot classifiers.
\newblock \emph{arXiv preprint arXiv:2303.15233}, 2023.

\bibitem[Croitoru et~al.(2023)Croitoru, Hondru, Ionescu, and Shah]{croitoru2023diffusion}
Florinel-Alin Croitoru, Vlad Hondru, Radu~Tudor Ionescu, and Mubarak Shah.
\newblock Diffusion models in vision: A survey.
\newblock \emph{IEEE Transactions on Pattern Analysis and Machine Intelligence}, 2023.

\bibitem[Efron(2011)]{efron2011tweedie}
Bradley Efron.
\newblock Tweedie’s formula and selection bias.
\newblock \emph{Journal of the American Statistical Association}, 106\penalty0 (496):\penalty0 1602--1614, 2011.

\bibitem[Esser et~al.(2023)Esser, Chiu, Atighehchian, Granskog, and Germanidis]{esser2023structure}
Patrick Esser, Johnathan Chiu, Parmida Atighehchian, Jonathan Granskog, and Anastasis Germanidis.
\newblock Structure and content-guided video synthesis with diffusion models.
\newblock In \emph{ICCV}, pages 7346--7356, 2023.

\bibitem[Gal et~al.(2022)Gal, Alaluf, Atzmon, Patashnik, Bermano, Chechik, and Cohen-Or]{gal2022image}
Rinon Gal, Yuval Alaluf, Yuval Atzmon, Or Patashnik, Amit~H Bermano, Gal Chechik, and Daniel Cohen-Or.
\newblock An image is worth one word: Personalizing text-to-image generation using textual inversion.
\newblock \emph{arXiv preprint arXiv:2208.01618}, 2022.

\bibitem[Ge et~al.(2023)Ge, Nah, Liu, Poon, Tao, Catanzaro, Jacobs, Huang, Liu, and Balaji]{ge2023preserve}
Songwei Ge, Seungjun Nah, Guilin Liu, Tyler Poon, Andrew Tao, Bryan Catanzaro, David Jacobs, Jia-Bin Huang, Ming-Yu Liu, and Yogesh Balaji.
\newblock Preserve your own correlation: A noise prior for video diffusion models.
\newblock In \emph{ICCV}, pages 22930--22941, 2023.

\bibitem[Gu et~al.(2023)Gu, Wang, Wu, Shi, Chen, Fan, Xiao, Zhao, Chang, Wu, et~al.]{gu2023mix}
Yuchao Gu, Xintao Wang, Jay~Zhangjie Wu, Yujun Shi, Yunpeng Chen, Zihan Fan, Wuyou Xiao, Rui Zhao, Shuning Chang, Weijia Wu, et~al.
\newblock Mix-of-show: Decentralized low-rank adaptation for multi-concept customization of diffusion models.
\newblock \emph{arXiv preprint arXiv:2305.18292}, 2023.

\bibitem[Harvey et~al.(2022)Harvey, Naderiparizi, Masrani, Weilbach, and Wood]{harvey2022flexible}
William Harvey, Saeid Naderiparizi, Vaden Masrani, Christian Weilbach, and Frank Wood.
\newblock Flexible diffusion modeling of long videos.
\newblock \emph{NeurIPS}, 35:\penalty0 27953--27965, 2022.

\bibitem[Ho et~al.(2020)Ho, Jain, and Abbeel]{ho2020denoising}
Jonathan Ho, Ajay Jain, and Pieter Abbeel.
\newblock Denoising diffusion probabilistic models.
\newblock \emph{NeurIPS}, 33:\penalty0 6840--6851, 2020.

\bibitem[Jafarian and Park(2021)]{jafarian2021learning}
Yasamin Jafarian and Hyun~Soo Park.
\newblock Learning high fidelity depths of dressed humans by watching social media dance videos.
\newblock In \emph{CVPR}, pages 12753--12762, 2021.

\bibitem[Karras et~al.(2023)Karras, Holynski, Wang, and Kemelmacher-Shlizerman]{karras2023dreampose}
Johanna Karras, Aleksander Holynski, Ting-Chun Wang, and Ira Kemelmacher-Shlizerman.
\newblock Dreampose: Fashion video synthesis with stable diffusion.
\newblock In \emph{ICCV}, pages 22680--22690, 2023.

\bibitem[Kawar et~al.(2023)Kawar, Zada, Lang, Tov, Chang, Dekel, Mosseri, and Irani]{kawar2023imagic}
Bahjat Kawar, Shiran Zada, Oran Lang, Omer Tov, Huiwen Chang, Tali Dekel, Inbar Mosseri, and Michal Irani.
\newblock Imagic: Text-based real image editing with diffusion models.
\newblock In \emph{CVPR}, pages 6007--6017, 2023.

\bibitem[Kirillov et~al.(2023)Kirillov, Mintun, Ravi, Mao, Rolland, Gustafson, Xiao, Whitehead, Berg, Lo, et~al.]{kirillov2023segment}
Alexander Kirillov, Eric Mintun, Nikhila Ravi, Hanzi Mao, Chloe Rolland, Laura Gustafson, Tete Xiao, Spencer Whitehead, Alexander~C Berg, Wan-Yen Lo, et~al.
\newblock Segment anything.
\newblock \emph{arXiv preprint arXiv:2304.02643}, 2023.

\bibitem[Kumari et~al.(2023)Kumari, Zhang, Zhang, Shechtman, and Zhu]{kumari2023multi}
Nupur Kumari, Bingliang Zhang, Richard Zhang, Eli Shechtman, and Jun-Yan Zhu.
\newblock Multi-concept customization of text-to-image diffusion.
\newblock In \emph{CVPR}, pages 1931--1941, 2023.

\bibitem[Liu et~al.(2023)Liu, Feng, Zhu, Zhang, Zheng, Liu, Zhao, Zhou, and Cao]{liu2023cones}
Zhiheng Liu, Ruili Feng, Kai Zhu, Yifei Zhang, Kecheng Zheng, Yu Liu, Deli Zhao, Jingren Zhou, and Yang Cao.
\newblock Cones: Concept neurons in diffusion models for customized generation.
\newblock \emph{arXiv preprint arXiv:2303.05125}, 2023.

\bibitem[Luo et~al.(2023)Luo, Chen, Zhang, Huang, Wang, Shen, Zhao, Zhou, and Tan]{luo2023videofusion}
Zhengxiong Luo, Dayou Chen, Yingya Zhang, Yan Huang, Liang Wang, Yujun Shen, Deli Zhao, Jingren Zhou, and Tieniu Tan.
\newblock Videofusion: Decomposed diffusion models for high-quality video generation.
\newblock In \emph{CVPR}, pages 10209--10218, 2023.

\bibitem[Mokady et~al.(2023)Mokady, Hertz, Aberman, Pritch, and Cohen-Or]{mokady2023null}
Ron Mokady, Amir Hertz, Kfir Aberman, Yael Pritch, and Daniel Cohen-Or.
\newblock Null-text inversion for editing real images using guided diffusion models.
\newblock In \emph{CVPR}, pages 6038--6047, 2023.

\bibitem[Nichol et~al.(2021)Nichol, Dhariwal, Ramesh, Shyam, Mishkin, McGrew, Sutskever, and Chen]{nichol2021glide}
Alex Nichol, Prafulla Dhariwal, Aditya Ramesh, Pranav Shyam, Pamela Mishkin, Bob McGrew, Ilya Sutskever, and Mark Chen.
\newblock Glide: Towards photorealistic image generation and editing with text-guided diffusion models.
\newblock \emph{arXiv preprint arXiv:2112.10741}, 2021.

\bibitem[Nichol and Dhariwal(2021)]{nichol2021improved}
Alexander~Quinn Nichol and Prafulla Dhariwal.
\newblock Improved denoising diffusion probabilistic models.
\newblock In \emph{ICML}, pages 8162--8171. PMLR, 2021.

\bibitem[Qi et~al.(2023)Qi, Cun, Zhang, Lei, Wang, Shan, and Chen]{qi2023fatezero}
Chenyang Qi, Xiaodong Cun, Yong Zhang, Chenyang Lei, Xintao Wang, Ying Shan, and Qifeng Chen.
\newblock Fatezero: Fusing attentions for zero-shot text-based video editing.
\newblock \emph{arXiv preprint arXiv:2303.09535}, 2023.

\bibitem[Radford et~al.(2021)Radford, Kim, Hallacy, Ramesh, Goh, Agarwal, Sastry, Askell, Mishkin, Clark, et~al.]{radford2021learning}
Alec Radford, Jong~Wook Kim, Chris Hallacy, Aditya Ramesh, Gabriel Goh, Sandhini Agarwal, Girish Sastry, Amanda Askell, Pamela Mishkin, Jack Clark, et~al.
\newblock Learning transferable visual models from natural language supervision.
\newblock In \emph{ICML}, pages 8748--8763, 2021.

\bibitem[Rombach et~al.(2022)Rombach, Blattmann, Lorenz, Esser, and Ommer]{rombach2022high}
Robin Rombach, Andreas Blattmann, Dominik Lorenz, Patrick Esser, and Bj{\"o}rn Ommer.
\newblock High-resolution image synthesis with latent diffusion models.
\newblock In \emph{CVPR}, pages 10684--10695, 2022.

\bibitem[Romera-Paredes and Torr(2015)]{romera2015embarrassingly}
Bernardino Romera-Paredes and Philip Torr.
\newblock An embarrassingly simple approach to zero-shot learning.
\newblock In \emph{ICML}, pages 2152--2161. PMLR, 2015.

\bibitem[Ruiz et~al.(2023)Ruiz, Li, Jampani, Pritch, Rubinstein, and Aberman]{ruiz2023dreambooth}
Nataniel Ruiz, Yuanzhen Li, Varun Jampani, Yael Pritch, Michael Rubinstein, and Kfir Aberman.
\newblock Dreambooth: Fine tuning text-to-image diffusion models for subject-driven generation.
\newblock In \emph{CVPR}, pages 22500--22510, 2023.

\bibitem[Saharia et~al.(2022)Saharia, Chan, Saxena, Li, Whang, Denton, Ghasemipour, Gontijo~Lopes, Karagol~Ayan, Salimans, et~al.]{saharia2022photorealistic}
Chitwan Saharia, William Chan, Saurabh Saxena, Lala Li, Jay Whang, Emily~L Denton, Kamyar Ghasemipour, Raphael Gontijo~Lopes, Burcu Karagol~Ayan, Tim Salimans, et~al.
\newblock Photorealistic text-to-image diffusion models with deep language understanding.
\newblock \emph{NeurIPS}, 35:\penalty0 36479--36494, 2022.

\bibitem[Siarohin et~al.(2019{\natexlab{a}})Siarohin, Lathuili{\`e}re, Tulyakov, Ricci, and Sebe]{siarohin2019animating}
Aliaksandr Siarohin, St{\'e}phane Lathuili{\`e}re, Sergey Tulyakov, Elisa Ricci, and Nicu Sebe.
\newblock Animating arbitrary objects via deep motion transfer.
\newblock In \emph{CVPR}, pages 2377--2386, 2019{\natexlab{a}}.

\bibitem[Siarohin et~al.(2019{\natexlab{b}})Siarohin, Lathuili{\`e}re, Tulyakov, Ricci, and Sebe]{siarohin2019first}
Aliaksandr Siarohin, St{\'e}phane Lathuili{\`e}re, Sergey Tulyakov, Elisa Ricci, and Nicu Sebe.
\newblock First order motion model for image animation.
\newblock \emph{NeurIPS}, 32, 2019{\natexlab{b}}.

\bibitem[Siarohin et~al.(2021)Siarohin, Woodford, Ren, Chai, and Tulyakov]{siarohin2021motion}
Aliaksandr Siarohin, Oliver~J Woodford, Jian Ren, Menglei Chai, and Sergey Tulyakov.
\newblock Motion representations for articulated animation.
\newblock In \emph{CVPR}, pages 13653--13662, 2021.

\bibitem[Song et~al.(2020)Song, Meng, and Ermon]{song2020denoising}
Jiaming Song, Chenlin Meng, and Stefano Ermon.
\newblock Denoising diffusion implicit models.
\newblock \emph{arXiv preprint arXiv:2010.02502}, 2020.

\bibitem[Tumanyan et~al.(2023)Tumanyan, Geyer, Bagon, and Dekel]{tumanyan2023plug}
Narek Tumanyan, Michal Geyer, Shai Bagon, and Tali Dekel.
\newblock Plug-and-play diffusion features for text-driven image-to-image translation.
\newblock In \emph{CVPR}, pages 1921--1930, 2023.

\bibitem[Unsplash()]{unsplash}
Unsplash.
\newblock https://unsplash.com/.

\bibitem[Wang et~al.(2023)Wang, Li, Lin, Lin, Yang, Zhang, Liu, and Wang]{wang2023disco}
Tan Wang, Linjie Li, Kevin Lin, Chung-Ching Lin, Zhengyuan Yang, Hanwang Zhang, Zicheng Liu, and Lijuan Wang.
\newblock Disco: Disentangled control for referring human dance generation in real world.
\newblock \emph{arXiv preprint arXiv:2307.00040}, 2023.

\bibitem[Wang et~al.(2018)Wang, Liu, Zhu, Liu, Tao, Kautz, and Catanzaro]{wang2018video}
Ting-Chun Wang, Ming-Yu Liu, Jun-Yan Zhu, Guilin Liu, Andrew Tao, Jan Kautz, and Bryan Catanzaro.
\newblock Video-to-video synthesis.
\newblock \emph{arXiv preprint arXiv:1808.06601}, 2018.

\bibitem[Wang et~al.(2019)Wang, Liu, Tao, Liu, Kautz, and Catanzaro]{wang2019few}
Ting-Chun Wang, Ming-Yu Liu, Andrew Tao, Guilin Liu, Jan Kautz, and Bryan Catanzaro.
\newblock Few-shot video-to-video synthesis.
\newblock \emph{arXiv preprint arXiv:1910.12713}, 2019.

\bibitem[Wu et~al.(2023)Wu, Ge, Wang, Lei, Gu, Shi, Hsu, Shan, Qie, and Shou]{wu2023tune}
Jay~Zhangjie Wu, Yixiao Ge, Xintao Wang, Stan~Weixian Lei, Yuchao Gu, Yufei Shi, Wynne Hsu, Ying Shan, Xiaohu Qie, and Mike~Zheng Shou.
\newblock Tune-a-video: One-shot tuning of image diffusion models for text-to-video generation.
\newblock In \emph{ICCV}, pages 7623--7633, 2023.

\bibitem[Xian et~al.(2017)Xian, Schiele, and Akata]{xian2017zero}
Yongqin Xian, Bernt Schiele, and Zeynep Akata.
\newblock Zero-shot learning-the good, the bad and the ugly.
\newblock In \emph{CVPR}, pages 4582--4591, 2017.

\bibitem[Yang et~al.(2023)Yang, Gu, Zhang, Zhang, Chen, Sun, Chen, and Wen]{yang2023paint}
Binxin Yang, Shuyang Gu, Bo Zhang, Ting Zhang, Xuejin Chen, Xiaoyan Sun, Dong Chen, and Fang Wen.
\newblock Paint by example: Exemplar-based image editing with diffusion models.
\newblock In \emph{CVPR}, pages 18381--18391, 2023.

\bibitem[Yu et~al.(2023)Yu, Sohn, Kim, and Shin]{yu2023video}
Sihyun Yu, Kihyuk Sohn, Subin Kim, and Jinwoo Shin.
\newblock Video probabilistic diffusion models in projected latent space.
\newblock In \emph{CVPR}, pages 18456--18466, 2023.

\bibitem[Zhang et~al.(2023{\natexlab{a}})Zhang, Rao, and Agrawala]{zhang2023adding}
Lvmin Zhang, Anyi Rao, and Maneesh Agrawala.
\newblock Adding conditional control to text-to-image diffusion models.
\newblock In \emph{ICCV}, pages 3836--3847, 2023{\natexlab{a}}.

\bibitem[Zhang et~al.(2023{\natexlab{b}})Zhang, Han, Ghosh, Metaxas, and Ren]{zhang2023sine}
Zhixing Zhang, Ligong Han, Arnab Ghosh, Dimitris~N Metaxas, and Jian Ren.
\newblock Sine: Single image editing with text-to-image diffusion models.
\newblock In \emph{CVPR}, pages 6027--6037, 2023{\natexlab{b}}.

\bibitem[Zhao and Zhang(2022)]{zhao2022thin}
Jian Zhao and Hui Zhang.
\newblock Thin-plate spline motion model for image animation.
\newblock In \emph{CVPR}, pages 3657--3666, 2022.

\end{thebibliography}


\begin{thebibliography}{16}
\providecommand{\natexlab}[1]{#1}
\providecommand{\url}[1]{\texttt{#1}}
\expandafter\ifx\csname urlstyle\endcsname\relax
  \providecommand{\doi}[1]{doi: #1}\else
  \providecommand{\doi}{doi: \begingroup \urlstyle{rm}\Url}\fi

\bibitem[Cao et~al.(2023)Cao, Wang, Qi, Shan, Qie, and Zheng]{cao2023masactrl}
Mingdeng Cao, Xintao Wang, Zhongang Qi, Ying Shan, Xiaohu Qie, and Yinqiang Zheng.
\newblock Masactrl: Tuning-free mutual self-attention control for consistent image synthesis and editing.
\newblock \emph{arXiv preprint arXiv:2304.08465}, 2023.

\bibitem[Cao et~al.(2017)Cao, Simon, Wei, and Sheikh]{cao2017realtime}
Zhe Cao, Tomas Simon, Shih-En Wei, and Yaser Sheikh.
\newblock Realtime multi-person 2d pose estimation using part affinity fields.
\newblock In \emph{CVPR}, pages 7291--7299, 2017.

\bibitem[Caron et~al.(2021)Caron, Touvron, Misra, J{\'e}gou, Mairal, Bojanowski, and Joulin]{caron2021emerging}
Mathilde Caron, Hugo Touvron, Ishan Misra, Herv{\'e} J{\'e}gou, Julien Mairal, Piotr Bojanowski, and Armand Joulin.
\newblock Emerging properties in self-supervised vision transformers.
\newblock In \emph{ICCV}, pages 9650--9660, 2021.

\bibitem[Gu et~al.(2023)Gu, Wang, Wu, Shi, Chen, Fan, Xiao, Zhao, Chang, Wu, et~al.]{gu2023mix}
Yuchao Gu, Xintao Wang, Jay~Zhangjie Wu, Yujun Shi, Yunpeng Chen, Zihan Fan, Wuyou Xiao, Rui Zhao, Shuning Chang, Weijia Wu, et~al.
\newblock Mix-of-show: Decentralized low-rank adaptation for multi-concept customization of diffusion models.
\newblock \emph{arXiv preprint arXiv:2305.18292}, 2023.

\bibitem[Jafarian and Park(2021)]{jafarian2021learning}
Yasamin Jafarian and Hyun~Soo Park.
\newblock Learning high fidelity depths of dressed humans by watching social media dance videos.
\newblock In \emph{CVPR}, pages 12753--12762, 2021.

\bibitem[Karras et~al.(2023)Karras, Holynski, Wang, and Kemelmacher-Shlizerman]{karras2023dreampose}
Johanna Karras, Aleksander Holynski, Ting-Chun Wang, and Ira Kemelmacher-Shlizerman.
\newblock Dreampose: Fashion video synthesis with stable diffusion.
\newblock In \emph{ICCV}, pages 22680--22690, 2023.

\bibitem[keypoint~evaluation metric()]{cocoeval}
MSCOCO keypoint~evaluation metric.
\newblock https://cocodataset.org/keypoints-eval.

\bibitem[Kirillov et~al.(2023)Kirillov, Mintun, Ravi, Mao, Rolland, Gustafson, Xiao, Whitehead, Berg, Lo, et~al.]{kirillov2023segment}
Alexander Kirillov, Eric Mintun, Nikhila Ravi, Hanzi Mao, Chloe Rolland, Laura Gustafson, Tete Xiao, Spencer Whitehead, Alexander~C Berg, Wan-Yen Lo, et~al.
\newblock Segment anything.
\newblock \emph{arXiv preprint arXiv:2304.02643}, 2023.

\bibitem[Mokady et~al.(2023)Mokady, Hertz, Aberman, Pritch, and Cohen-Or]{mokady2023null}
Ron Mokady, Amir Hertz, Kfir Aberman, Yael Pritch, and Daniel Cohen-Or.
\newblock Null-text inversion for editing real images using guided diffusion models.
\newblock In \emph{CVPR}, pages 6038--6047, 2023.

\bibitem[Qi et~al.(2023)Qi, Cun, Zhang, Lei, Wang, Shan, and Chen]{qi2023fatezero}
Chenyang Qi, Xiaodong Cun, Yong Zhang, Chenyang Lei, Xintao Wang, Ying Shan, and Qifeng Chen.
\newblock Fatezero: Fusing attentions for zero-shot text-based video editing.
\newblock \emph{arXiv preprint arXiv:2303.09535}, 2023.

\bibitem[Radford et~al.(2021)Radford, Kim, Hallacy, Ramesh, Goh, Agarwal, Sastry, Askell, Mishkin, Clark, et~al.]{radford2021learning}
Alec Radford, Jong~Wook Kim, Chris Hallacy, Aditya Ramesh, Gabriel Goh, Sandhini Agarwal, Girish Sastry, Amanda Askell, Pamela Mishkin, Jack Clark, et~al.
\newblock Learning transferable visual models from natural language supervision.
\newblock In \emph{ICML}, pages 8748--8763, 2021.

\bibitem[Romera-Paredes and Torr(2015)]{romera2015embarrassingly}
Bernardino Romera-Paredes and Philip Torr.
\newblock An embarrassingly simple approach to zero-shot learning.
\newblock In \emph{ICML}, pages 2152--2161. PMLR, 2015.

\bibitem[Ruiz et~al.(2023)Ruiz, Li, Jampani, Pritch, Rubinstein, and Aberman]{ruiz2023dreambooth}
Nataniel Ruiz, Yuanzhen Li, Varun Jampani, Yael Pritch, Michael Rubinstein, and Kfir Aberman.
\newblock Dreambooth: Fine tuning text-to-image diffusion models for subject-driven generation.
\newblock In \emph{CVPR}, pages 22500--22510, 2023.

\bibitem[Unsplash()]{unsplash}
Unsplash.
\newblock https://unsplash.com/.

\bibitem[Wang et~al.(2023)Wang, Li, Lin, Lin, Yang, Zhang, Liu, and Wang]{wang2023disco}
Tan Wang, Linjie Li, Kevin Lin, Chung-Ching Lin, Zhengyuan Yang, Hanwang Zhang, Zicheng Liu, and Lijuan Wang.
\newblock Disco: Disentangled control for referring human dance generation in real world.
\newblock \emph{arXiv preprint arXiv:2307.00040}, 2023.

\bibitem[Zhang et~al.(2023)Zhang, Rao, and Agrawala]{zhang2023adding}
Lvmin Zhang, Anyi Rao, and Maneesh Agrawala.
\newblock Adding conditional control to text-to-image diffusion models.
\newblock In \emph{ICCV}, pages 3836--3847, 2023.

\end{thebibliography}
}

\end{document}


\maketitle

\section{Dataset}

\begin{figure*}[]
\centering
\includegraphics[width=1\textwidth]{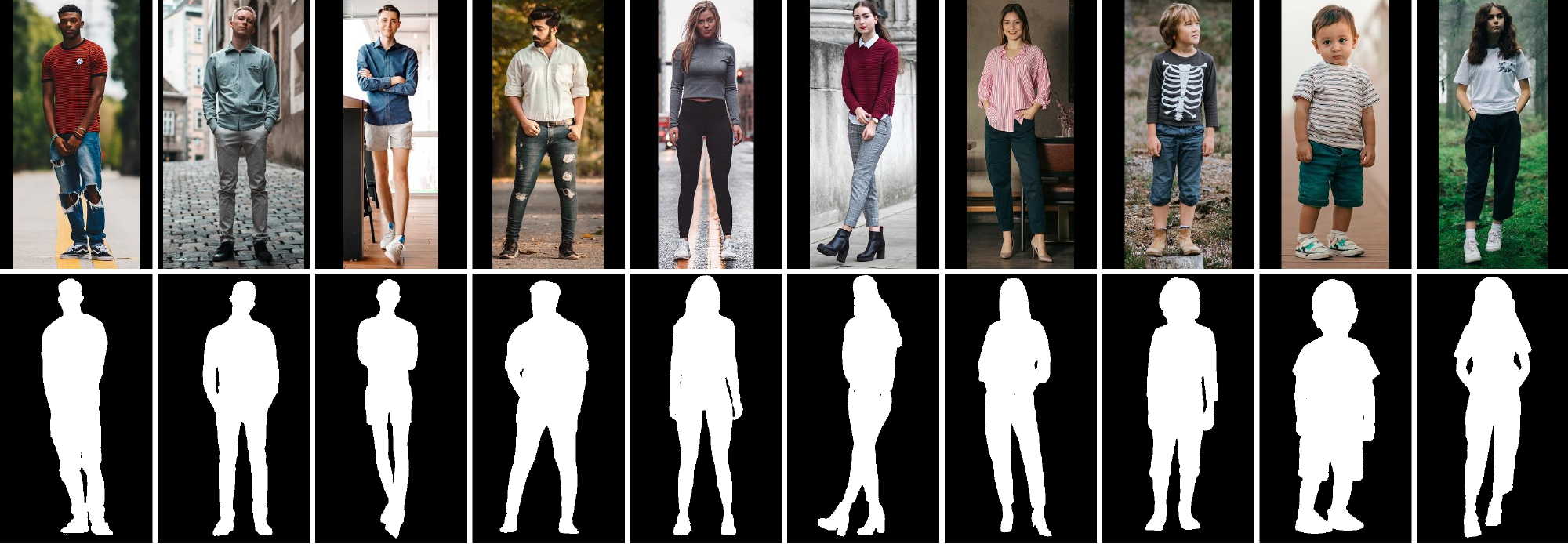} 
\caption{Images of persons and their masks in our proposed dataset.}
\label{fig:dataset_fgs}
\end{figure*}

\begin{figure*}[]
\centering
\includegraphics[width=1\textwidth]{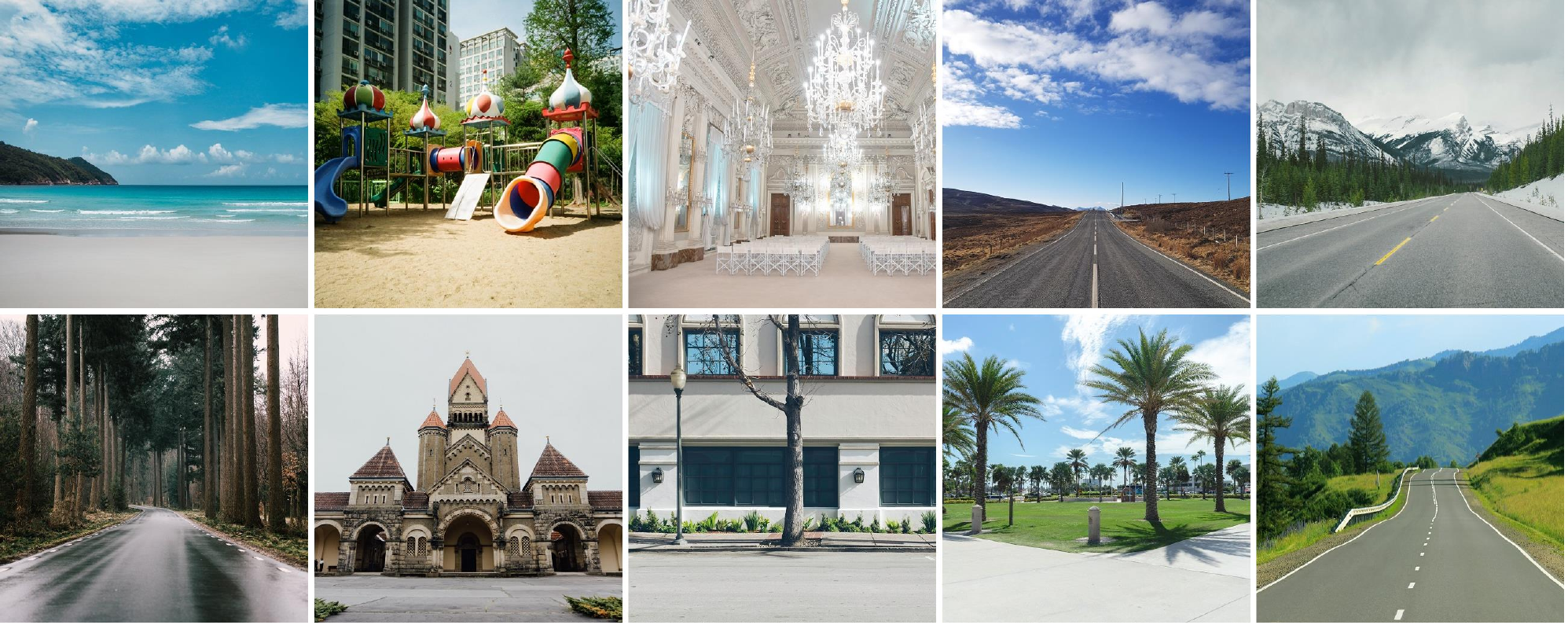} 
\caption{Images of backgrounds in our proposed dataset.}
\label{fig:dataset_bgs}
\end{figure*}

\begin{figure*}[]
\centering
\includegraphics[width=1\textwidth]{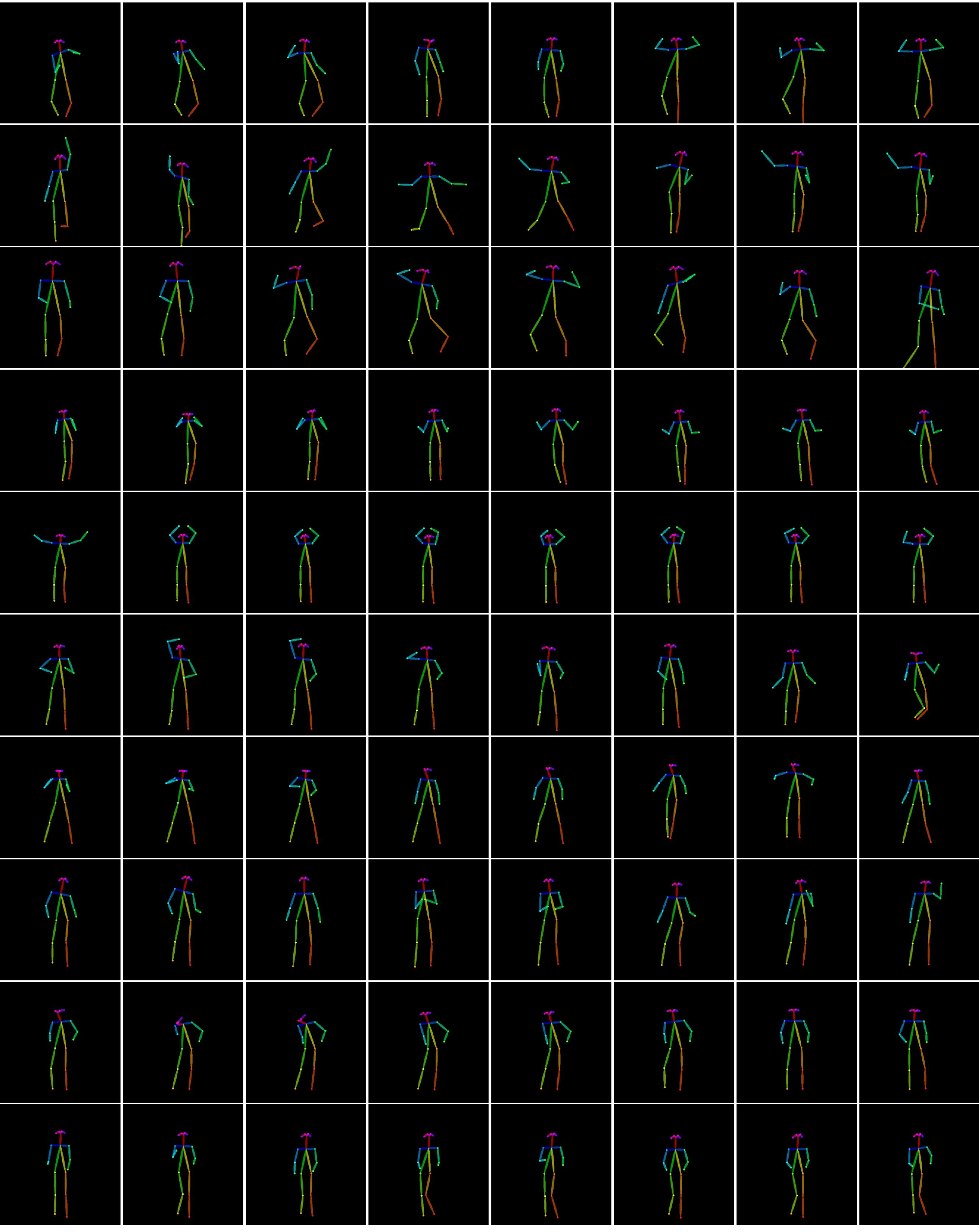} 
\caption{Pose squences in our proposed dataset.}
\label{fig:dataset_poses}
\end{figure*}

\indent\hspace{1em} Our dataset for compositional human dance generation (cHDG) consists of 10 persons, 10 backgrounds, and 10 pose sequences, which can be composed to synthesize thousands of different videos. Images of the persons and backgrounds are presented in Fig.\ref{fig:dataset_fgs} and Fig.\ref{fig:dataset_bgs}, respectively, which were sourced from Unsplash \cite{unsplash}. The human masks in Fig.\ref{fig:dataset_fgs} used for composing multi-person images are computed by SAM \cite{kirillov2023segment}. We also present the poses sequences for our dataset in Fig.\ref{fig:dataset_poses}, which are computed via OpenPose \cite{cao2017realtime} using videos from the public TikTok dataset \cite{jafarian2021learning}. Multi-person poses are obtained by composing corresponding single poses. We will make our dataset publicly available.

\section{Evaluation Metrics}
\indent\hspace{1em} Since there is no previous works on zero-shot cHDG, we propose to utilize the following metrics for evaluation borrowed from related tasks \cite{qi2023fatezero,ruiz2023dreambooth,cao2017realtime,romera2015embarrassingly}: 
\begin{itemize}
\item `CLIP-I': the average cosine similarities between the composed reference image and each frame via a pretrained CLIP \cite{radford2021learning} model. We use \href{https://huggingface.co/openai/clip-vit-base-patch32}{CLIP-ViT-b32} in our evaluation.
\item `DINO': the average pairwise cosine similarity between the DINO \cite{caron2021emerging} embeddings of generated frames and those of the composed reference image. We use \href{https://huggingface.co/facebook/dino-vitb16}{DINO-ViT-b16 }in our evaluation. As discussed in \cite{ruiz2023dreambooth}, `DINO' is a better metric than `CLIP-I' for temporal consistency because DINO is trained under self-supervised training objective to encourage distinction of
unique features of a image. In contrast, CLIP is trained with paired image-text data to ignore the differences between subjects of the same class.
\item `mAP': we take the input pose sequences as ground truth and compute the mean average precision of generated videos to measure the pose accuracy. To be specific, we follow MSCOCO evaluation \cite{cocoeval} to compute the object keypoint similarity (OKS) and obtain the mean average precision over 10 OKS thresholds:
\begin{equation}
    \operatorname{OKS}=\cfrac{\sum_i \operatorname{exp}(-d^2_i/2s^2k^2_i)\delta(v_i>0)}{\sum_i \delta(v_{i}>0)},
\end{equation}
where $d_i$ are the Eucildean distances between each corresponding ground truth and detected keypoint and $v_i$ are the visibility flags of the ground truth. $s$ is the object scale and $k_i$ is a per-keypont constant that controls falloff.
\item `H': harmonic mean value of DINO score and mAP, namely:
\begin{equation}
    \operatorname{H} = \cfrac{2\times \operatorname{DINO} \times \operatorname{mAP}}{\operatorname{DINO} + \operatorname{mAP}}.
\end{equation}
H is our preferred metric since it can simultaneously measure the temporal consistency and pose accuracy.

\end{itemize}

\begin{table*}[]
\centering
\begin{tabular}{lccccccc}
\hline\noalign{\smallskip}
Method  &Zero-Shot &CLIP-I$\uparrow$ &DINO$\uparrow$  &mAP$\uparrow$ &H$\uparrow$ &User-Study$\downarrow$  &Storage\\
\hline\noalign{\smallskip}
DreamPose  & \XSolidBrush&\textbf{0.87} &\textbf{0.89} &0.46 &0.61&5.60 &8.5GB\\
DisCO  & \XSolidBrush &0.78 &0.76 &0.87 &\textcolor{blue}{0.81}&\textcolor{blue}{2.77} &4.3GB\\
DDIM\&ControlNet &\Checkmark &0.70 &0.64 &0.83 &0.72 &4.27 &-\\
Null\&ControlNet &\Checkmark &0.69 &0.61 &0.87 &0.72&4.16 &11.3MB\\
MasaCtrl &\Checkmark &0.65 &0.51 &\textcolor{blue}{0.93} &0.66&5.27 &-\\
Mix-of-Show &\Checkmark &0.77& 0.65 &\textbf{0.97} &0.78& 4.87&2.0GB\\
\hline\noalign{\smallskip}
Ours &\Checkmark &\textcolor{blue}{0.83}&\textcolor{blue}{0.83}&0.91 &\textbf{0.87}&\textbf{1.23}&22.6MB\\
\hline\noalign{\smallskip}
\end{tabular}
\caption{\textbf{Quantitative results.} We compare our method wit six state-of-the-art baselines. The best and the second best methods are shown in \textbf{bold} and \textcolor{blue}{blue}, respectively.}
\label{tab:comparison}
\end{table*}

\section{Additional Experiments}

\subsection{Additional Comparisons}

\indent\hspace{1em} We compare our method with the following state-of-the-art baselines on cHDG: 
\begin{itemize}
    \item DreamPose \cite{karras2023dreampose} proposes a diffusion-based method for generating fashion videos from still images, where a dual CLIP-VAE image encoder and an adapter module are employed to replace the text encoder in the original Stable Diffusion architecture. Meanwhile, a two-phase fine-tuning scheme is introduced to improve the image fidelity and temporal consistency. Following the official implementation of \href{https://github.com/johannakarras/DreamPose}{DreamPose}, we finetune the DreamPose model trained on the UBC Fashion dataset using the reference image to create a subject-specific model. 
    \item DisCo \cite{wang2023disco} presents a model architecture with disentangled foreground-background-pose control and human attribute pre-training to improve the fidelity of generation. To synthesize videos with target poses, we separate the foreground and background from the reference image and fed them to the official pretrained \href{https://github.com/Wangt-CN/DisCO}{DisCO} model.
    \item DDIM\&ControlNet uses DDIM inversion \cite{mokady2023null} to invert the reference image into the latent space and generates videos via pretrained pose-conditioned ControlNet \cite{zhang2023adding}.
    \item Null\&ControlNet first uses null-text inversion \cite{mokady2023null} to invert the reference image into the latent space and optimize a set of null-text embeddings, then generates videos via pretrained pose-conditioned ControlNet.
    \item MasaCtrl \cite{cao2023masactrl} performs complex non-rigid frame-wise image editing via a mutual self-attention mechanism. We follow the official implementation of \href{https://github.com/TencentARC/MasaCtrl}{MasaCtrl} to obtain the results.
    \item Mix-of-Show \cite{gu2023mix} adopts an embedding-decomposed LoRA for single concept tuning and gradient fusion for center-node concept fusion. Following the official implementation of \href{https://github.com/TencentARC/Mix-of-Show}{Mix-of-Show}, we first perform single-client concept tuning for each person and background. Then, we fuse multiple ED-LoRAs to generate videos via regionally controllable multi-concept sampling.
\end{itemize}

Among the compared baselines, DreamPose and DisCO are supervised methods that require large numbers of video data to train their models. In contrast, DDIM\&ControlNet, Null\&ControlNet, MasaCtrl, Mix-of-Show, and our method are zero-shot approaches with no video data utilized.

\begin{figure*}[]
\centering
\includegraphics[width=1\textwidth]{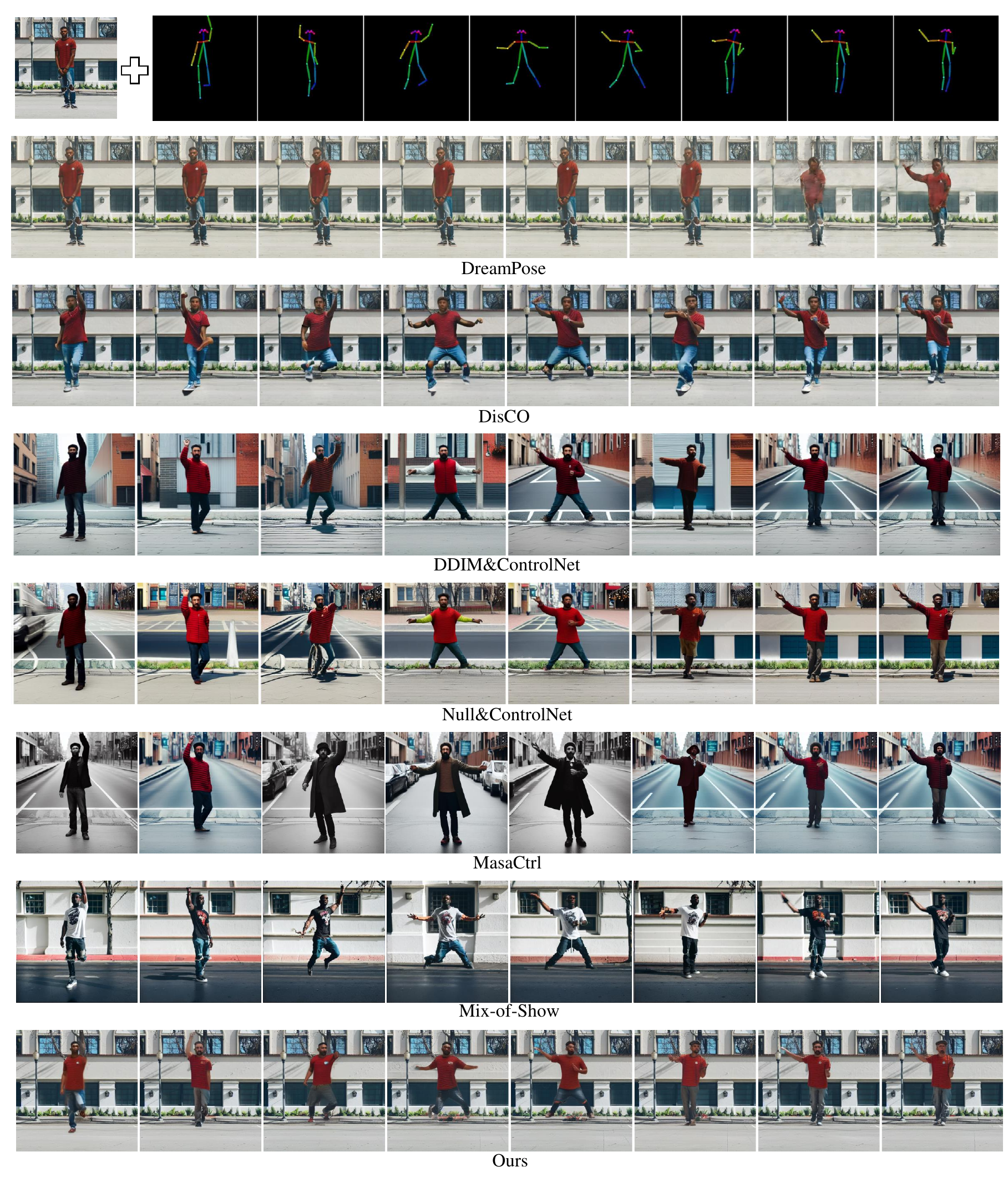} 
\caption{\textbf{Single-person results.} We present quantitative results with single person of our method and six state-of-the-art baselines on cHDG. The composed reference image and driving poses are shown in the top row.}
\label{fig:comparison_1}
\end{figure*}

\begin{figure*}[]
\centering
\includegraphics[width=1\textwidth]{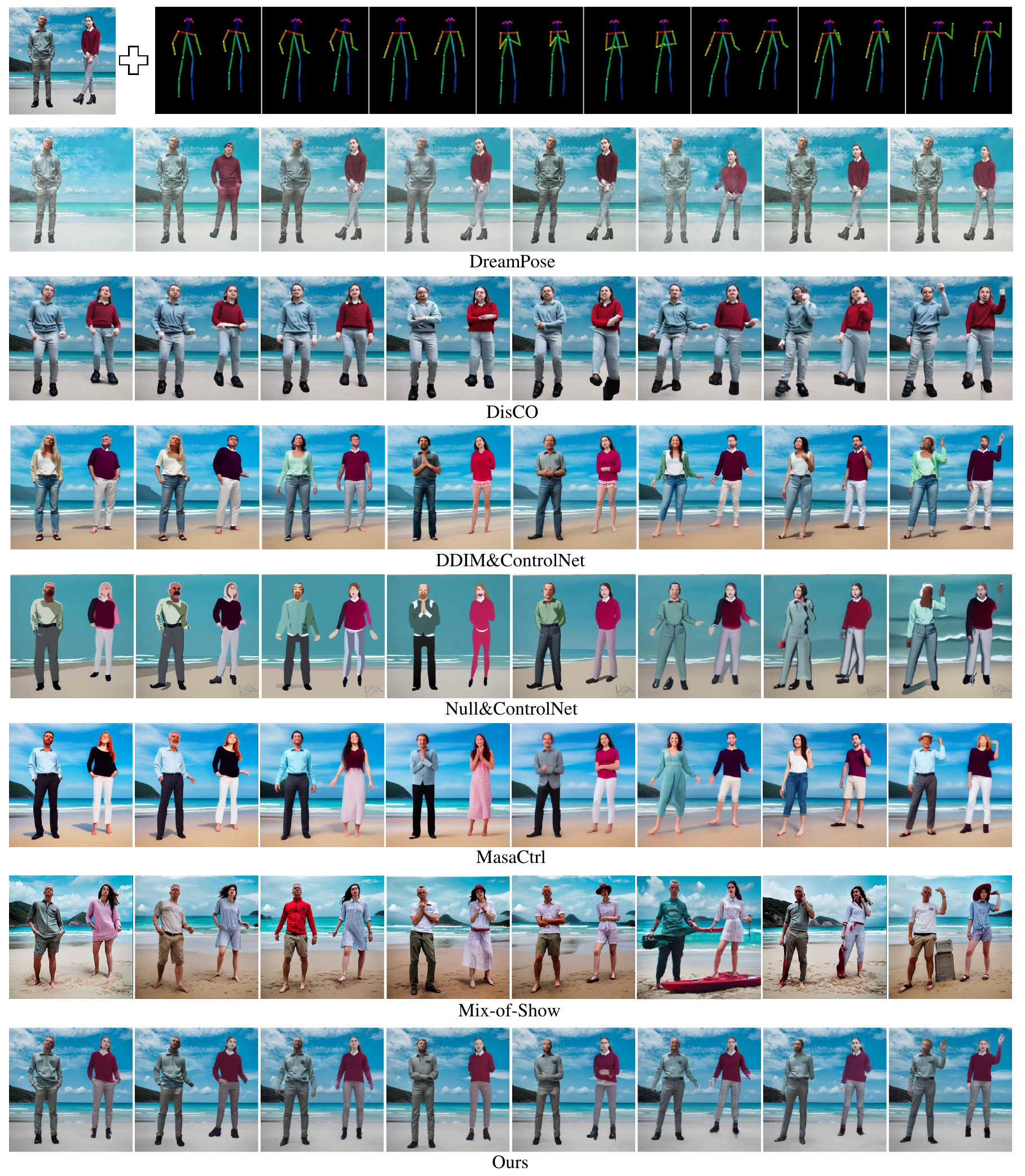} 
\caption{\textbf{Two-person results.}  We present quantitative results with two persons of our method and six state-of-the-art baselines on cHDG. The composed reference image and driving poses are shown in the top row.}
\vspace{1cm}
\label{fig:comparison_2}
\end{figure*}

\begin{figure*}[]
\centering
\includegraphics[width=1\textwidth]{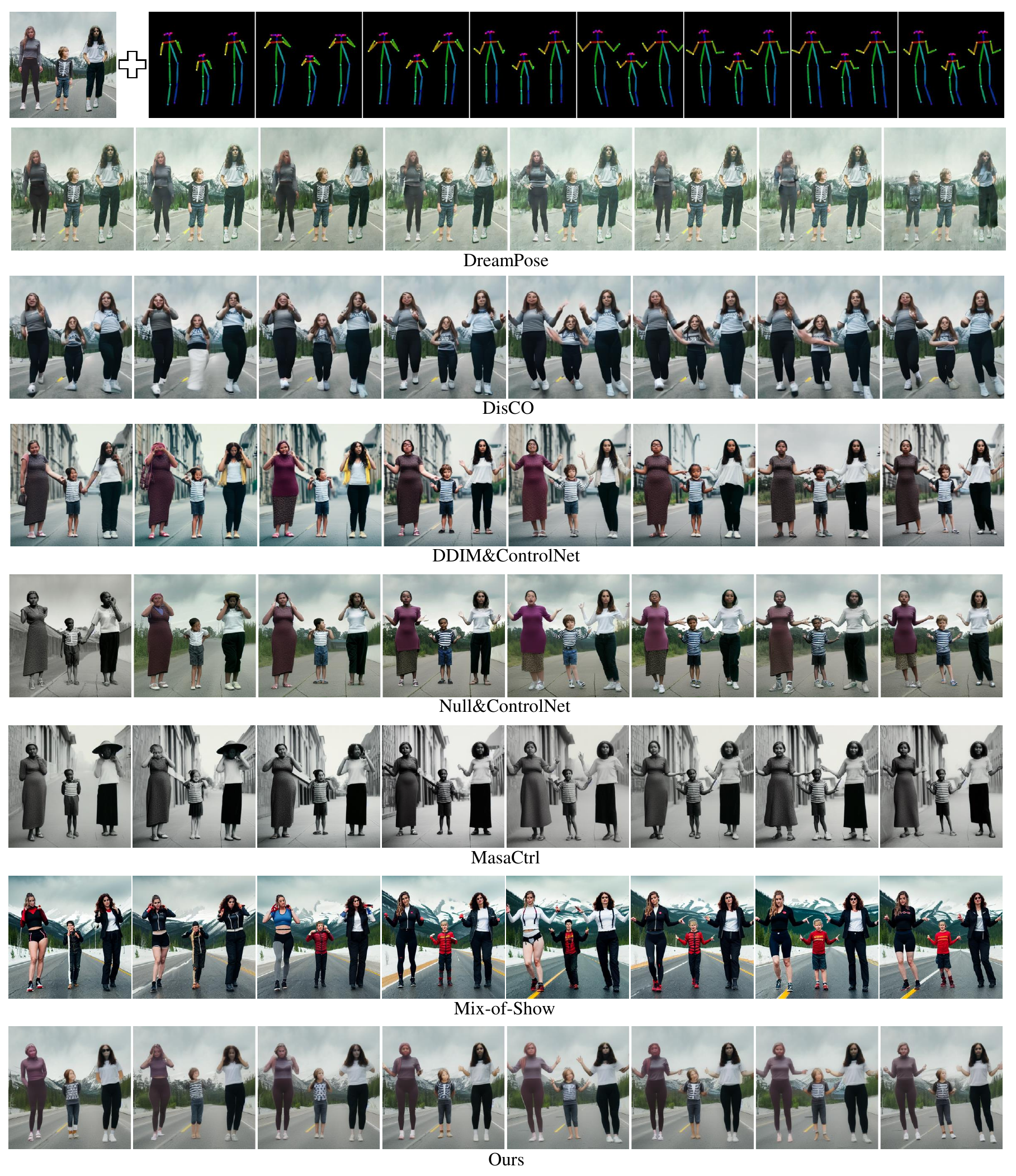} 
\caption{\textbf{Three-person results.}  We present quantitative results with three persons of our method and six state-of-the-art baselines on cHDG. The composed reference image and driving poses are shown in the top row.}
\label{fig:comparison_3}
\end{figure*}

We present the quantitative results in Tab.\ref{tab:comparison}. As indicated by the harmonic mean value H, our method achieves the best temporal consistency and pose accuracy against baselines.
Fig.\ref{fig:comparison_1}, Fig.\ref{fig:comparison_2}, and Fig.\ref{fig:comparison_3}  show the qualitative results with different number of persons.  Since DreamPose is trained on fashion videos with empty background and easy catwalk poses, it can not synthesize video in real-world scenarios with complex backgrounds and poses. DisCO, another well-trained supervised method, performs much better than DreamPose due to its disentangled foreground-background-pose control. However, the appearances of generated videos, especially the multi-person videos, are still inconsistent with the reference image. Benefited from the pretrained pose-conditioned ControlNet, zero-shot baselines, \emph{i.e.}, DDIM\&ControlNet, Null\&ControlNet, MasaCtrl, and Mix-of-Show, can synthesize videos that precisely follow the driving poses. But their appearances are far from inconsistent with the reference image due to the generative prior within pretrained DMs. Our method can synthesize a video that simultaneously retains the appearance consistent with the reference images and precisely follows the poses.

\subsection{User Study}
We also conduct a user study to evaluate the overall generation quality, where we ask 20 persons to rank different methods by considering the frame-wise image fidelity, temporal consistency and pose accuracy of generated videos. As shown in Tab.\ref{tab:comparison}, our method perform best among all the compared ones.

\subsection{Storage}
We report the storage costs of different methods in Tab.\ref{tab:comparison}. We only need to store the optimized generalizable text embeddings, while DreamPose, DisCO, and Mix-of-Show need to store their whole models.

\subsection{Additional Ablation Study}
\begin{figure*}[]
\centering
\includegraphics[width=1\textwidth]{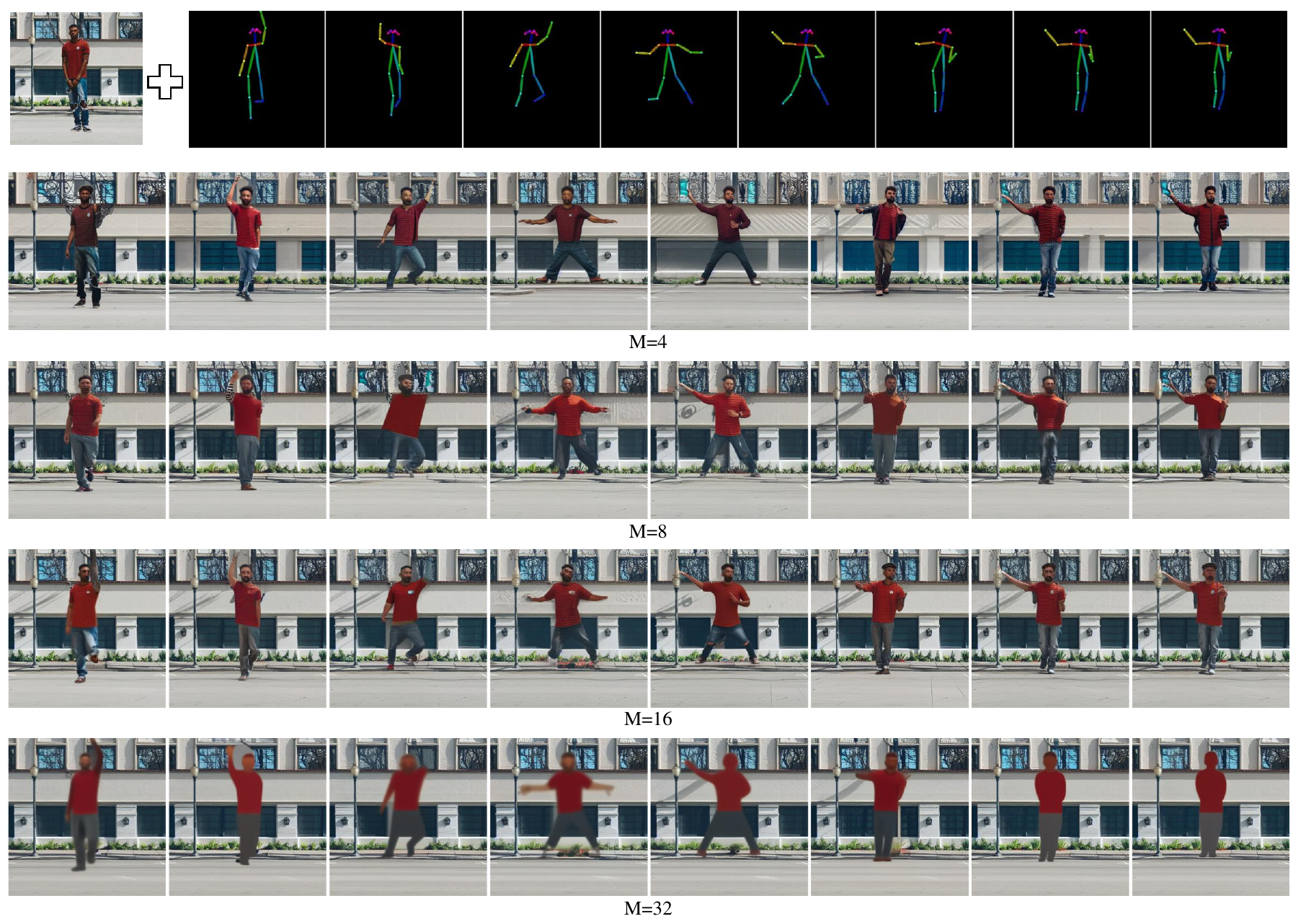} 
\caption{\textbf{Ablation study on $M$.}  We present quantitative results with different number of augmented images. The composed reference image and driving poses are shown in the top row.}
\label{fig:ablation_m}
\end{figure*}

\textbf{Influence of the number of augmented images.} In this experiment, we perform an ablation study to investigate the influence of the number of augmented images $M$ for generalizable text embeddings optimization. As shown in Fig.\ref{fig:ablation_m}, the quality of synthesized videos is clearly improved as $M$ increases from 4 to 16. When we further set $M$ to 32, the optimized text embeddings seem to focus more on the unchanged background and perform worse on the foreground consistency.

{
    \small
    \bibliographystyle{ieeenat_fullname}
    \bibliography{supp}
}